\documentclass[10pt, table, xcdraw]{article}

\usepackage[accepted]{tmlr}

\usepackage[utf8]{inputenc} %
\usepackage[T1]{fontenc}    %
\usepackage{hyperref}       %
\usepackage{url}            %
\usepackage{booktabs}       %
\usepackage{amsfonts}       %
\usepackage{nicefrac}       %
\usepackage{microtype}      %
\usepackage{wrapfig}

\usepackage{researchpack}

\usepackage[capitalize,nameinlink]{cleveref}
\usepackage{subcaption}
\usepackage{multirow}
\usepackage{comment}

\hypersetup{
    colorlinks=true,
    citecolor=blue,
    linkcolor=blue,
    filecolor=blue,
    urlcolor=blue,
}

\definecolor {snow}                {rgb}{1.00,0.98,0.98}
\definecolor {ghostwhite}          {rgb}{0.97,0.97,1.00}
\definecolor {whitesmoke}          {rgb}{0.96,0.96,0.96}
\definecolor {gainsboro}           {rgb}{0.86,0.86,0.86}
\definecolor {floralwhite}         {rgb}{1.00,0.98,0.94}
\definecolor {oldlace}             {rgb}{0.99,0.96,0.90}
\definecolor {linen}               {rgb}{0.98,0.94,0.90}
\definecolor {antiquewhite}        {rgb}{0.98,0.92,0.84}
\definecolor {papayawhip}          {rgb}{1.00,0.94,0.84}
\definecolor {blanchedalmond}      {rgb}{1.00,0.92,0.80}
\definecolor {bisque}              {rgb}{1.00,0.89,0.77}
\definecolor {peachpuff}           {rgb}{1.00,0.85,0.73}
\definecolor {navajowhite}         {rgb}{1.00,0.87,0.68}
\definecolor {moccasin}            {rgb}{1.00,0.89,0.71}
\definecolor {cornsilk}            {rgb}{1.00,0.97,0.86}
\definecolor {ivory}               {rgb}{1.00,1.00,0.94}
\definecolor {lemonchiffon}        {rgb}{1.00,0.98,0.80}
\definecolor {seashell}            {rgb}{1.00,0.96,0.93}
\definecolor {honeydew}            {rgb}{0.94,1.00,0.94}
\definecolor {mintcream}           {rgb}{0.96,1.00,0.98}
\definecolor {azure}               {rgb}{0.94,1.00,1.00}
\definecolor {aliceblue}           {rgb}{0.94,0.97,1.00}
\definecolor {lavender}            {rgb}{0.90,0.90,0.98}
\definecolor {lavenderblush}       {rgb}{1.00,0.94,0.96}
\definecolor {mistyrose}           {rgb}{1.00,0.89,0.88}
\definecolor {white}               {rgb}{1.00,1.00,1.00}
\definecolor {black}               {rgb}{0.00,0.00,0.00}
\definecolor {darkslategray}       {rgb}{0.18,0.31,0.31}
\definecolor {dimgray}             {rgb}{0.41,0.41,0.41}
\definecolor {slategray}           {rgb}{0.44,0.50,0.56}
\definecolor {lightslategray}      {rgb}{0.47,0.53,0.60}
\definecolor {gray}                {rgb}{0.75,0.75,0.75}
\definecolor {lightgrey}           {rgb}{0.83,0.83,0.83}
\definecolor {midnightblue}        {rgb}{0.10,0.10,0.44}
\definecolor {navy}                {rgb}{0.00,0.00,0.50}
\definecolor {cornflowerblue}      {rgb}{0.39,0.58,0.93}
\definecolor {darkslateblue}       {rgb}{0.28,0.24,0.55}
\definecolor {slateblue}           {rgb}{0.42,0.35,0.80}
\definecolor {mediumslateblue}     {rgb}{0.48,0.41,0.93}
\definecolor {lightslateblue}      {rgb}{0.52,0.44,1.00}
\definecolor {mediumblue}          {rgb}{0.00,0.00,0.80}
\definecolor {royalblue}           {rgb}{0.25,0.41,0.88}
\definecolor {blue}                {rgb}{0.00,0.00,1.00}
\definecolor {dodgerblue}          {rgb}{0.12,0.56,1.00}
\definecolor {deepskyblue}         {rgb}{0.00,0.75,1.00}
\definecolor {skyblue}             {rgb}{0.53,0.81,0.92}
\definecolor {lightskyblue}        {rgb}{0.53,0.81,0.98}
\definecolor {steelblue}           {rgb}{0.27,0.51,0.71}
\definecolor {lightsteelblue}      {rgb}{0.69,0.77,0.87}
\definecolor {lightblue}           {rgb}{0.68,0.85,0.90}
\definecolor {powderblue}          {rgb}{0.69,0.88,0.90}
\definecolor {paleturquoise}       {rgb}{0.69,0.93,0.93}
\definecolor {darkturquoise}       {rgb}{0.00,0.81,0.82}
\definecolor {mediumturquoise}     {rgb}{0.28,0.82,0.80}
\definecolor {turquoise}           {rgb}{0.25,0.88,0.82}
\definecolor {cyan}                {rgb}{0.00,1.00,1.00}
\definecolor {lightcyan}           {rgb}{0.88,1.00,1.00}
\definecolor {cadetblue}           {rgb}{0.37,0.62,0.63}
\definecolor {mediumaquamarine}    {rgb}{0.40,0.80,0.67}
\definecolor {aquamarine}          {rgb}{0.50,1.00,0.83}
\definecolor {darkgreen}           {rgb}{0.00,0.39,0.00}
\definecolor {darkolivegreen}      {rgb}{0.33,0.42,0.18}
\definecolor {darkseagreen}        {rgb}{0.56,0.74,0.56}
\definecolor {seagreen}            {rgb}{0.18,0.55,0.34}
\definecolor {mediumseagreen}      {rgb}{0.24,0.70,0.44}
\definecolor {lightseagreen}       {rgb}{0.13,0.70,0.67}
\definecolor {palegreen}           {rgb}{0.60,0.98,0.60}
\definecolor {springgreen}         {rgb}{0.00,1.00,0.50}
\definecolor {lawngreen}           {rgb}{0.49,0.99,0.00}
\definecolor {green}               {rgb}{0.00,1.00,0.00}
\definecolor {chartreuse}          {rgb}{0.50,1.00,0.00}
\definecolor {mediumspringgreen}   {rgb}{0.00,0.98,0.60}
\definecolor {greenyellow}         {rgb}{0.68,1.00,0.18}
\definecolor {limegreen}           {rgb}{0.20,0.80,0.20}
\definecolor {yellowgreen}         {rgb}{0.60,0.80,0.20}
\definecolor {forestgreen}         {rgb}{0.13,0.55,0.13}
\definecolor {olivedrab}           {rgb}{0.42,0.56,0.14}
\definecolor {darkkhaki}           {rgb}{0.74,0.72,0.42}
\definecolor {khaki}               {rgb}{0.94,0.90,0.55}
\definecolor {palegoldenrod}       {rgb}{0.93,0.91,0.67}
\definecolor {lightgoldenrodyellow} {rgb}{0.98,0.98,0.82}
\definecolor {lightyellow}         {rgb}{1.00,1.00,0.88}
\definecolor {yellow}              {rgb}{1.00,1.00,0.00}
\definecolor {gold}                {rgb}{1.00,0.84,0.00}
\definecolor {lightgoldenrod}      {rgb}{0.93,0.87,0.51}
\definecolor {goldenrod}           {rgb}{0.85,0.65,0.13}
\definecolor {darkgoldenrod}       {rgb}{0.72,0.53,0.04}
\definecolor {rosybrown}           {rgb}{0.74,0.56,0.56}
\definecolor {indianred}           {rgb}{0.80,0.36,0.36}
\definecolor {saddlebrown}         {rgb}{0.55,0.27,0.07}
\definecolor {sienna}              {rgb}{0.63,0.32,0.18}
\definecolor {peru}                {rgb}{0.80,0.52,0.25}
\definecolor {burlywood}           {rgb}{0.87,0.72,0.53}
\definecolor {beige}               {rgb}{0.96,0.96,0.86}
\definecolor {wheat}               {rgb}{0.96,0.87,0.70}
\definecolor {sandybrown}          {rgb}{0.96,0.64,0.38}
\definecolor {tan}                 {rgb}{0.82,0.71,0.55}
\definecolor {chocolate}           {rgb}{0.82,0.41,0.12}
\definecolor {firebrick}           {rgb}{0.70,0.13,0.13}
\definecolor {brown}               {rgb}{0.65,0.16,0.16}
\definecolor {darksalmon}          {rgb}{0.91,0.59,0.48}
\definecolor {salmon}              {rgb}{0.98,0.50,0.45}
\definecolor {lightsalmon}         {rgb}{1.00,0.63,0.48}
\definecolor {orange}              {rgb}{1.00,0.65,0.00}
\definecolor {darkorange}          {rgb}{1.00,0.55,0.00}
\definecolor {coral}               {rgb}{1.00,0.50,0.31}
\definecolor {lightcoral}          {rgb}{0.94,0.50,0.50}
\definecolor {tomato}              {rgb}{1.00,0.39,0.28}
\definecolor {orangered}           {rgb}{1.00,0.27,0.00}
\definecolor {red}                 {rgb}{1.00,0.00,0.00}
\definecolor {hotpink}             {rgb}{1.00,0.41,0.71}
\definecolor {deeppink}            {rgb}{1.00,0.08,0.58}
\definecolor {pink}                {rgb}{1.00,0.75,0.80}
\definecolor {lightpink}           {rgb}{1.00,0.71,0.76}
\definecolor {palevioletred}       {rgb}{0.86,0.44,0.58}
\definecolor {maroon}              {rgb}{0.69,0.19,0.38}
\definecolor {mediumvioletred}     {rgb}{0.78,0.08,0.52}
\definecolor {violetred}           {rgb}{0.82,0.13,0.56}
\definecolor {magenta}             {rgb}{1.00,0.00,1.00}
\definecolor {violet}              {rgb}{0.93,0.51,0.93}
\definecolor {plum}                {rgb}{0.87,0.63,0.87}
\definecolor {orchid}              {rgb}{0.85,0.44,0.84}
\definecolor {mediumorchid}        {rgb}{0.73,0.33,0.83}
\definecolor {darkorchid}          {rgb}{0.60,0.20,0.80}
\definecolor {darkviolet}          {rgb}{0.58,0.00,0.83}
\definecolor {blueviolet}          {rgb}{0.54,0.17,0.89}
\definecolor {purple}              {rgb}{0.63,0.13,0.94}
\definecolor {mediumpurple}        {rgb}{0.58,0.44,0.86}
\definecolor {thistle}             {rgb}{0.85,0.75,0.85}
\definecolor {snow2}               {rgb}{0.93,0.91,0.91}
\definecolor {snow3}               {rgb}{0.80,0.79,0.79}
\definecolor {snow4}               {rgb}{0.55,0.54,0.54}
\definecolor {seashell2}           {rgb}{0.93,0.90,0.87}
\definecolor {seashell3}           {rgb}{0.80,0.77,0.75}
\definecolor {seashell4}           {rgb}{0.55,0.53,0.51}
\definecolor {antiquewhite1}       {rgb}{1.00,0.94,0.86}
\definecolor {antiquewhite2}       {rgb}{0.93,0.87,0.80}
\definecolor {antiquewhite3}       {rgb}{0.80,0.75,0.69}
\definecolor {antiquewhite4}       {rgb}{0.55,0.51,0.47}
\definecolor {bisque2}             {rgb}{0.93,0.84,0.72}
\definecolor {bisque3}             {rgb}{0.80,0.72,0.62}
\definecolor {bisque4}             {rgb}{0.55,0.49,0.42}
\definecolor {peachpuff2}          {rgb}{0.93,0.80,0.68}
\definecolor {peachpuff3}          {rgb}{0.80,0.69,0.58}
\definecolor {peachpuff4}          {rgb}{0.55,0.47,0.40}
\definecolor {navajowhite2}        {rgb}{0.93,0.81,0.63}
\definecolor {navajowhite3}        {rgb}{0.80,0.70,0.55}
\definecolor {navajowhite4}        {rgb}{0.55,0.47,0.37}
\definecolor {lemonchiffon2}       {rgb}{0.93,0.91,0.75}
\definecolor {lemonchiffon3}       {rgb}{0.80,0.79,0.65}
\definecolor {lemonchiffon4}       {rgb}{0.55,0.54,0.44}
\definecolor {cornsilk2}           {rgb}{0.93,0.91,0.80}
\definecolor {cornsilk3}           {rgb}{0.80,0.78,0.69}
\definecolor {cornsilk4}           {rgb}{0.55,0.53,0.47}
\definecolor {ivory2}              {rgb}{0.93,0.93,0.88}
\definecolor {ivory3}              {rgb}{0.80,0.80,0.76}
\definecolor {ivory4}              {rgb}{0.55,0.55,0.51}
\definecolor {honeydew2}           {rgb}{0.88,0.93,0.88}
\definecolor {honeydew3}           {rgb}{0.76,0.80,0.76}
\definecolor {honeydew4}           {rgb}{0.51,0.55,0.51}
\definecolor {lavenderblush2}      {rgb}{0.93,0.88,0.90}
\definecolor {lavenderblush3}      {rgb}{0.80,0.76,0.77}
\definecolor {lavenderblush4}      {rgb}{0.55,0.51,0.53}
\definecolor {mistyrose2}          {rgb}{0.93,0.84,0.82}
\definecolor {mistyrose3}          {rgb}{0.80,0.72,0.71}
\definecolor {mistyrose4}          {rgb}{0.55,0.49,0.48}
\definecolor {azure2}              {rgb}{0.88,0.93,0.93}
\definecolor {azure3}              {rgb}{0.76,0.80,0.80}
\definecolor {azure4}              {rgb}{0.51,0.55,0.55}
\definecolor {slateblue1}          {rgb}{0.51,0.44,1.00}
\definecolor {slateblue2}          {rgb}{0.48,0.40,0.93}
\definecolor {slateblue3}          {rgb}{0.41,0.35,0.80}
\definecolor {slateblue4}          {rgb}{0.28,0.24,0.55}
\definecolor {royalblue1}          {rgb}{0.28,0.46,1.00}
\definecolor {royalblue2}          {rgb}{0.26,0.43,0.93}
\definecolor {royalblue3}          {rgb}{0.23,0.37,0.80}
\definecolor {royalblue4}          {rgb}{0.15,0.25,0.55}
\definecolor {blue2}               {rgb}{0.00,0.00,0.93}
\definecolor {blue4}               {rgb}{0.00,0.00,0.55}
\definecolor {dodgerblue2}         {rgb}{0.11,0.53,0.93}
\definecolor {dodgerblue3}         {rgb}{0.09,0.45,0.80}
\definecolor {dodgerblue4}         {rgb}{0.06,0.31,0.55}
\definecolor {steelblue1}          {rgb}{0.39,0.72,1.00}
\definecolor {steelblue2}          {rgb}{0.36,0.67,0.93}
\definecolor {steelblue3}          {rgb}{0.31,0.58,0.80}
\definecolor {steelblue4}          {rgb}{0.21,0.39,0.55}
\definecolor {deepskyblue2}        {rgb}{0.00,0.70,0.93}
\definecolor {deepskyblue3}        {rgb}{0.00,0.60,0.80}
\definecolor {deepskyblue4}        {rgb}{0.00,0.41,0.55}
\definecolor {skyblue1}            {rgb}{0.53,0.81,1.00}
\definecolor {skyblue2}            {rgb}{0.49,0.75,0.93}
\definecolor {skyblue3}            {rgb}{0.42,0.65,0.80}
\definecolor {skyblue4}            {rgb}{0.29,0.44,0.55}
\definecolor {lightskyblue1}       {rgb}{0.69,0.89,1.00}
\definecolor {lightskyblue2}       {rgb}{0.64,0.83,0.93}
\definecolor {lightskyblue3}       {rgb}{0.55,0.71,0.80}
\definecolor {lightskyblue4}       {rgb}{0.38,0.48,0.55}
\definecolor {slategray1}          {rgb}{0.78,0.89,1.00}
\definecolor {slategray2}          {rgb}{0.73,0.83,0.93}
\definecolor {slategray3}          {rgb}{0.62,0.71,0.80}
\definecolor {slategray4}          {rgb}{0.42,0.48,0.55}
\definecolor {lightsteelblue1}     {rgb}{0.79,0.88,1.00}
\definecolor {lightsteelblue2}     {rgb}{0.74,0.82,0.93}
\definecolor {lightsteelblue3}     {rgb}{0.64,0.71,0.80}
\definecolor {lightsteelblue4}     {rgb}{0.43,0.48,0.55}
\definecolor {lightblue1}          {rgb}{0.75,0.94,1.00}
\definecolor {lightblue2}          {rgb}{0.70,0.87,0.93}
\definecolor {lightblue3}          {rgb}{0.60,0.75,0.80}
\definecolor {lightblue4}          {rgb}{0.41,0.51,0.55}
\definecolor {lightcyan2}          {rgb}{0.82,0.93,0.93}
\definecolor {lightcyan3}          {rgb}{0.71,0.80,0.80}
\definecolor {lightcyan4}          {rgb}{0.48,0.55,0.55}
\definecolor {paleturquoise1}      {rgb}{0.73,1.00,1.00}
\definecolor {paleturquoise2}      {rgb}{0.68,0.93,0.93}
\definecolor {paleturquoise3}      {rgb}{0.59,0.80,0.80}
\definecolor {paleturquoise4}      {rgb}{0.40,0.55,0.55}
\definecolor {cadetblue1}          {rgb}{0.60,0.96,1.00}
\definecolor {cadetblue2}          {rgb}{0.56,0.90,0.93}
\definecolor {cadetblue3}          {rgb}{0.48,0.77,0.80}
\definecolor {cadetblue4}          {rgb}{0.33,0.53,0.55}
\definecolor {turquoise1}          {rgb}{0.00,0.96,1.00}
\definecolor {turquoise2}          {rgb}{0.00,0.90,0.93}
\definecolor {turquoise3}          {rgb}{0.00,0.77,0.80}
\definecolor {turquoise4}          {rgb}{0.00,0.53,0.55}
\definecolor {cyan2}               {rgb}{0.00,0.93,0.93}
\definecolor {cyan3}               {rgb}{0.00,0.80,0.80}
\definecolor {cyan4}               {rgb}{0.00,0.55,0.55}
\definecolor {darkslategray1}      {rgb}{0.59,1.00,1.00}
\definecolor {darkslategray2}      {rgb}{0.55,0.93,0.93}
\definecolor {darkslategray3}      {rgb}{0.47,0.80,0.80}
\definecolor {darkslategray4}      {rgb}{0.32,0.55,0.55}
\definecolor {aquamarine2}         {rgb}{0.46,0.93,0.78}
\definecolor {aquamarine4}         {rgb}{0.27,0.55,0.45}
\definecolor {darkseagreen1}       {rgb}{0.76,1.00,0.76}
\definecolor {darkseagreen2}       {rgb}{0.71,0.93,0.71}
\definecolor {darkseagreen3}       {rgb}{0.61,0.80,0.61}
\definecolor {darkseagreen4}       {rgb}{0.41,0.55,0.41}
\definecolor {seagreen1}           {rgb}{0.33,1.00,0.62}
\definecolor {seagreen2}           {rgb}{0.31,0.93,0.58}
\definecolor {seagreen3}           {rgb}{0.26,0.80,0.50}
\definecolor {palegreen1}          {rgb}{0.60,1.00,0.60}
\definecolor {palegreen2}          {rgb}{0.56,0.93,0.56}
\definecolor {palegreen3}          {rgb}{0.49,0.80,0.49}
\definecolor {palegreen4}          {rgb}{0.33,0.55,0.33}
\definecolor {springgreen2}        {rgb}{0.00,0.93,0.46}
\definecolor {springgreen3}        {rgb}{0.00,0.80,0.40}
\definecolor {springgreen4}        {rgb}{0.00,0.55,0.27}
\definecolor {green2}              {rgb}{0.00,0.93,0.00}
\definecolor {green3}              {rgb}{0.00,0.80,0.00}
\definecolor {green4}              {rgb}{0.00,0.55,0.00}
\definecolor {chartreuse2}         {rgb}{0.46,0.93,0.00}
\definecolor {chartreuse3}         {rgb}{0.40,0.80,0.00}
\definecolor {chartreuse4}         {rgb}{0.27,0.55,0.00}
\definecolor {olivedrab1}          {rgb}{0.75,1.00,0.24}
\definecolor {olivedrab2}          {rgb}{0.70,0.93,0.23}
\definecolor {olivedrab4}          {rgb}{0.41,0.55,0.13}
\definecolor {darkolivegreen1}     {rgb}{0.79,1.00,0.44}
\definecolor {darkolivegreen2}     {rgb}{0.74,0.93,0.41}
\definecolor {darkolivegreen3}     {rgb}{0.64,0.80,0.35}
\definecolor {darkolivegreen4}     {rgb}{0.43,0.55,0.24}
\definecolor {khaki1}              {rgb}{1.00,0.96,0.56}
\definecolor {khaki2}              {rgb}{0.93,0.90,0.52}
\definecolor {khaki3}              {rgb}{0.80,0.78,0.45}
\definecolor {khaki4}              {rgb}{0.55,0.53,0.31}
\definecolor {lightgoldenrod1}     {rgb}{1.00,0.93,0.55}
\definecolor {lightgoldenrod2}     {rgb}{0.93,0.86,0.51}
\definecolor {lightgoldenrod3}     {rgb}{0.80,0.75,0.44}
\definecolor {lightgoldenrod4}     {rgb}{0.55,0.51,0.30}
\definecolor {lightyellow2}        {rgb}{0.93,0.93,0.82}
\definecolor {lightyellow3}        {rgb}{0.80,0.80,0.71}
\definecolor {lightyellow4}        {rgb}{0.55,0.55,0.48}
\definecolor {yellow2}             {rgb}{0.93,0.93,0.00}
\definecolor {yellow3}             {rgb}{0.80,0.80,0.00}
\definecolor {yellow4}             {rgb}{0.55,0.55,0.00}
\definecolor {gold2}               {rgb}{0.93,0.79,0.00}
\definecolor {gold3}               {rgb}{0.80,0.68,0.00}
\definecolor {gold4}               {rgb}{0.55,0.46,0.00}
\definecolor {goldenrod1}          {rgb}{1.00,0.76,0.15}
\definecolor {goldenrod2}          {rgb}{0.93,0.71,0.13}
\definecolor {goldenrod3}          {rgb}{0.80,0.61,0.11}
\definecolor {goldenrod4}          {rgb}{0.55,0.41,0.08}
\definecolor {darkgoldenrod1}      {rgb}{1.00,0.73,0.06}
\definecolor {darkgoldenrod2}      {rgb}{0.93,0.68,0.05}
\definecolor {darkgoldenrod3}      {rgb}{0.80,0.58,0.05}
\definecolor {darkgoldenrod4}      {rgb}{0.55,0.40,0.03}
\definecolor {rosybrown1}          {rgb}{1.00,0.76,0.76}
\definecolor {rosybrown2}          {rgb}{0.93,0.71,0.71}
\definecolor {rosybrown3}          {rgb}{0.80,0.61,0.61}
\definecolor {rosybrown4}          {rgb}{0.55,0.41,0.41}
\definecolor {indianred1}          {rgb}{1.00,0.42,0.42}
\definecolor {indianred2}          {rgb}{0.93,0.39,0.39}
\definecolor {indianred3}          {rgb}{0.80,0.33,0.33}
\definecolor {indianred4}          {rgb}{0.55,0.23,0.23}
\definecolor {sienna1}             {rgb}{1.00,0.51,0.28}
\definecolor {sienna2}             {rgb}{0.93,0.47,0.26}
\definecolor {sienna3}             {rgb}{0.80,0.41,0.22}
\definecolor {sienna4}             {rgb}{0.55,0.28,0.15}
\definecolor {burlywood1}          {rgb}{1.00,0.83,0.61}
\definecolor {burlywood2}          {rgb}{0.93,0.77,0.57}
\definecolor {burlywood3}          {rgb}{0.80,0.67,0.49}
\definecolor {burlywood4}          {rgb}{0.55,0.45,0.33}
\definecolor {wheat1}              {rgb}{1.00,0.91,0.73}
\definecolor {wheat2}              {rgb}{0.93,0.85,0.68}
\definecolor {wheat3}              {rgb}{0.80,0.73,0.59}
\definecolor {wheat4}              {rgb}{0.55,0.49,0.40}
\definecolor {tan1}                {rgb}{1.00,0.65,0.31}
\definecolor {tan2}                {rgb}{0.93,0.60,0.29}
\definecolor {tan4}                {rgb}{0.55,0.35,0.17}
\definecolor {chocolate1}          {rgb}{1.00,0.50,0.14}
\definecolor {chocolate2}          {rgb}{0.93,0.46,0.13}
\definecolor {chocolate3}          {rgb}{0.80,0.40,0.11}
\definecolor {firebrick1}          {rgb}{1.00,0.19,0.19}
\definecolor {firebrick2}          {rgb}{0.93,0.17,0.17}
\definecolor {firebrick3}          {rgb}{0.80,0.15,0.15}
\definecolor {firebrick4}          {rgb}{0.55,0.10,0.10}
\definecolor {brown1}              {rgb}{1.00,0.25,0.25}
\definecolor {brown2}              {rgb}{0.93,0.23,0.23}
\definecolor {brown3}              {rgb}{0.80,0.20,0.20}
\definecolor {brown4}              {rgb}{0.55,0.14,0.14}
\definecolor {salmon1}             {rgb}{1.00,0.55,0.41}
\definecolor {salmon2}             {rgb}{0.93,0.51,0.38}
\definecolor {salmon3}             {rgb}{0.80,0.44,0.33}
\definecolor {salmon4}             {rgb}{0.55,0.30,0.22}
\definecolor {lightsalmon2}        {rgb}{0.93,0.58,0.45}
\definecolor {lightsalmon3}        {rgb}{0.80,0.51,0.38}
\definecolor {lightsalmon4}        {rgb}{0.55,0.34,0.26}
\definecolor {orange2}             {rgb}{0.93,0.60,0.00}
\definecolor {orange3}             {rgb}{0.80,0.52,0.00}
\definecolor {orange4}             {rgb}{0.55,0.35,0.00}
\definecolor {darkorange1}         {rgb}{1.00,0.50,0.00}
\definecolor {darkorange2}         {rgb}{0.93,0.46,0.00}
\definecolor {darkorange3}         {rgb}{0.80,0.40,0.00}
\definecolor {darkorange4}         {rgb}{0.55,0.27,0.00}
\definecolor {coral1}              {rgb}{1.00,0.45,0.34}
\definecolor {coral2}              {rgb}{0.93,0.42,0.31}
\definecolor {coral3}              {rgb}{0.80,0.36,0.27}
\definecolor {coral4}              {rgb}{0.55,0.24,0.18}
\definecolor {tomato2}             {rgb}{0.93,0.36,0.26}
\definecolor {tomato3}             {rgb}{0.80,0.31,0.22}
\definecolor {tomato4}             {rgb}{0.55,0.21,0.15}
\definecolor {orangered2}          {rgb}{0.93,0.25,0.00}
\definecolor {orangered3}          {rgb}{0.80,0.22,0.00}
\definecolor {orangered4}          {rgb}{0.55,0.15,0.00}
\definecolor {red2}                {rgb}{0.93,0.00,0.00}
\definecolor {red3}                {rgb}{0.80,0.00,0.00}
\definecolor {red4}                {rgb}{0.55,0.00,0.00}
\definecolor {deeppink2}           {rgb}{0.93,0.07,0.54}
\definecolor {deeppink3}           {rgb}{0.80,0.06,0.46}
\definecolor {deeppink4}           {rgb}{0.55,0.04,0.31}
\definecolor {hotpink1}            {rgb}{1.00,0.43,0.71}
\definecolor {hotpink2}            {rgb}{0.93,0.42,0.65}
\definecolor {hotpink3}            {rgb}{0.80,0.38,0.56}
\definecolor {hotpink4}            {rgb}{0.55,0.23,0.38}
\definecolor {pink1}               {rgb}{1.00,0.71,0.77}
\definecolor {pink2}               {rgb}{0.93,0.66,0.72}
\definecolor {pink3}               {rgb}{0.80,0.57,0.62}
\definecolor {pink4}               {rgb}{0.55,0.39,0.42}
\definecolor {lightpink1}          {rgb}{1.00,0.68,0.73}
\definecolor {lightpink2}          {rgb}{0.93,0.64,0.68}
\definecolor {lightpink3}          {rgb}{0.80,0.55,0.58}
\definecolor {lightpink4}          {rgb}{0.55,0.37,0.40}
\definecolor {palevioletred1}      {rgb}{1.00,0.51,0.67}
\definecolor {palevioletred2}      {rgb}{0.93,0.47,0.62}
\definecolor {palevioletred3}      {rgb}{0.80,0.41,0.54}
\definecolor {palevioletred4}      {rgb}{0.55,0.28,0.36}
\definecolor {maroon1}             {rgb}{1.00,0.20,0.70}
\definecolor {maroon2}             {rgb}{0.93,0.19,0.65}
\definecolor {maroon3}             {rgb}{0.80,0.16,0.56}
\definecolor {maroon4}             {rgb}{0.55,0.11,0.38}
\definecolor {violetred1}          {rgb}{1.00,0.24,0.59}
\definecolor {violetred2}          {rgb}{0.93,0.23,0.55}
\definecolor {violetred3}          {rgb}{0.80,0.20,0.47}
\definecolor {violetred4}          {rgb}{0.55,0.13,0.32}
\definecolor {magenta2}            {rgb}{0.93,0.00,0.93}
\definecolor {magenta3}            {rgb}{0.80,0.00,0.80}
\definecolor {magenta4}            {rgb}{0.55,0.00,0.55}
\definecolor {orchid1}             {rgb}{1.00,0.51,0.98}
\definecolor {orchid2}             {rgb}{0.93,0.48,0.91}
\definecolor {orchid3}             {rgb}{0.80,0.41,0.79}
\definecolor {orchid4}             {rgb}{0.55,0.28,0.54}
\definecolor {plum1}               {rgb}{1.00,0.73,1.00}
\definecolor {plum2}               {rgb}{0.93,0.68,0.93}
\definecolor {plum3}               {rgb}{0.80,0.59,0.80}
\definecolor {plum4}               {rgb}{0.55,0.40,0.55}
\definecolor {mediumorchid1}       {rgb}{0.88,0.40,1.00}
\definecolor {mediumorchid2}       {rgb}{0.82,0.37,0.93}
\definecolor {mediumorchid3}       {rgb}{0.71,0.32,0.80}
\definecolor {mediumorchid4}       {rgb}{0.48,0.22,0.55}
\definecolor {darkorchid1}         {rgb}{0.75,0.24,1.00}
\definecolor {darkorchid2}         {rgb}{0.70,0.23,0.93}
\definecolor {darkorchid3}         {rgb}{0.60,0.20,0.80}
\definecolor {darkorchid4}         {rgb}{0.41,0.13,0.55}
\definecolor {purple1}             {rgb}{0.61,0.19,1.00}
\definecolor {purple2}             {rgb}{0.57,0.17,0.93}
\definecolor {purple3}             {rgb}{0.49,0.15,0.80}
\definecolor {purple4}             {rgb}{0.33,0.10,0.55}
\definecolor {mediumpurple1}       {rgb}{0.67,0.51,1.00}
\definecolor {mediumpurple2}       {rgb}{0.62,0.47,0.93}
\definecolor {mediumpurple3}       {rgb}{0.54,0.41,0.80}
\definecolor {mediumpurple4}       {rgb}{0.36,0.28,0.55}
\definecolor {thistle1}            {rgb}{1.00,0.88,1.00}
\definecolor {thistle2}            {rgb}{0.93,0.82,0.93}
\definecolor {thistle3}            {rgb}{0.80,0.71,0.80}
\definecolor {thistle4}            {rgb}{0.55,0.48,0.55}
\definecolor {gray1}               {rgb}{0.01,0.01,0.01}
\definecolor {gray2}               {rgb}{0.02,0.02,0.02}
\definecolor {gray3}               {rgb}{0.03,0.03,0.03}
\definecolor {gray4}               {rgb}{0.04,0.04,0.04}
\definecolor {gray5}               {rgb}{0.05,0.05,0.05}
\definecolor {gray6}               {rgb}{0.06,0.06,0.06}
\definecolor {gray7}               {rgb}{0.07,0.07,0.07}
\definecolor {gray8}               {rgb}{0.08,0.08,0.08}
\definecolor {gray9}               {rgb}{0.09,0.09,0.09}
\definecolor {gray10}              {rgb}{0.10,0.10,0.10}
\definecolor {gray11}              {rgb}{0.11,0.11,0.11}
\definecolor {gray12}              {rgb}{0.12,0.12,0.12}
\definecolor {gray13}              {rgb}{0.13,0.13,0.13}
\definecolor {gray14}              {rgb}{0.14,0.14,0.14}
\definecolor {gray15}              {rgb}{0.15,0.15,0.15}
\definecolor {gray16}              {rgb}{0.16,0.16,0.16}
\definecolor {gray17}              {rgb}{0.17,0.17,0.17}
\definecolor {gray18}              {rgb}{0.18,0.18,0.18}
\definecolor {gray19}              {rgb}{0.19,0.19,0.19}
\definecolor {gray20}              {rgb}{0.20,0.20,0.20}
\definecolor {gray21}              {rgb}{0.21,0.21,0.21}
\definecolor {gray22}              {rgb}{0.22,0.22,0.22}
\definecolor {gray23}              {rgb}{0.23,0.23,0.23}
\definecolor {gray24}              {rgb}{0.24,0.24,0.24}
\definecolor {gray25}              {rgb}{0.25,0.25,0.25}
\definecolor {gray26}              {rgb}{0.26,0.26,0.26}
\definecolor {gray27}              {rgb}{0.27,0.27,0.27}
\definecolor {gray28}              {rgb}{0.28,0.28,0.28}
\definecolor {gray29}              {rgb}{0.29,0.29,0.29}
\definecolor {gray30}              {rgb}{0.30,0.30,0.30}
\definecolor {gray31}              {rgb}{0.31,0.31,0.31}
\definecolor {gray32}              {rgb}{0.32,0.32,0.32}
\definecolor {gray33}              {rgb}{0.33,0.33,0.33}
\definecolor {gray34}              {rgb}{0.34,0.34,0.34}
\definecolor {gray35}              {rgb}{0.35,0.35,0.35}
\definecolor {gray36}              {rgb}{0.36,0.36,0.36}
\definecolor {gray37}              {rgb}{0.37,0.37,0.37}
\definecolor {gray38}              {rgb}{0.38,0.38,0.38}
\definecolor {gray39}              {rgb}{0.39,0.39,0.39}
\definecolor {gray40}              {rgb}{0.40,0.40,0.40}
\definecolor {gray42}              {rgb}{0.42,0.42,0.42}
\definecolor {gray43}              {rgb}{0.43,0.43,0.43}
\definecolor {gray44}              {rgb}{0.44,0.44,0.44}
\definecolor {gray45}              {rgb}{0.45,0.45,0.45}
\definecolor {gray46}              {rgb}{0.46,0.46,0.46}
\definecolor {gray47}              {rgb}{0.47,0.47,0.47}
\definecolor {gray48}              {rgb}{0.48,0.48,0.48}
\definecolor {gray49}              {rgb}{0.49,0.49,0.49}
\definecolor {gray50}              {rgb}{0.50,0.50,0.50}
\definecolor {gray51}              {rgb}{0.51,0.51,0.51}
\definecolor {gray52}              {rgb}{0.52,0.52,0.52}
\definecolor {gray53}              {rgb}{0.53,0.53,0.53}
\definecolor {gray54}              {rgb}{0.54,0.54,0.54}
\definecolor {gray55}              {rgb}{0.55,0.55,0.55}
\definecolor {gray56}              {rgb}{0.56,0.56,0.56}
\definecolor {gray57}              {rgb}{0.57,0.57,0.57}
\definecolor {gray58}              {rgb}{0.58,0.58,0.58}
\definecolor {gray59}              {rgb}{0.59,0.59,0.59}
\definecolor {gray60}              {rgb}{0.60,0.60,0.60}
\definecolor {gray61}              {rgb}{0.61,0.61,0.61}
\definecolor {gray62}              {rgb}{0.62,0.62,0.62}
\definecolor {gray63}              {rgb}{0.63,0.63,0.63}
\definecolor {gray64}              {rgb}{0.64,0.64,0.64}
\definecolor {gray65}              {rgb}{0.65,0.65,0.65}
\definecolor {gray66}              {rgb}{0.66,0.66,0.66}
\definecolor {gray67}              {rgb}{0.67,0.67,0.67}
\definecolor {gray68}              {rgb}{0.68,0.68,0.68}
\definecolor {gray69}              {rgb}{0.69,0.69,0.69}
\definecolor {gray70}              {rgb}{0.70,0.70,0.70}
\definecolor {gray71}              {rgb}{0.71,0.71,0.71}
\definecolor {gray72}              {rgb}{0.72,0.72,0.72}
\definecolor {gray73}              {rgb}{0.73,0.73,0.73}
\definecolor {gray74}              {rgb}{0.74,0.74,0.74}
\definecolor {gray75}              {rgb}{0.75,0.75,0.75}
\definecolor {gray76}              {rgb}{0.76,0.76,0.76}
\definecolor {gray77}              {rgb}{0.77,0.77,0.77}
\definecolor {gray78}              {rgb}{0.78,0.78,0.78}
\definecolor {gray79}              {rgb}{0.79,0.79,0.79}
\definecolor {gray80}              {rgb}{0.80,0.80,0.80}
\definecolor {gray81}              {rgb}{0.81,0.81,0.81}
\definecolor {gray82}              {rgb}{0.82,0.82,0.82}
\definecolor {gray83}              {rgb}{0.83,0.83,0.83}
\definecolor {gray84}              {rgb}{0.84,0.84,0.84}
\definecolor {gray85}              {rgb}{0.85,0.85,0.85}
\definecolor {gray86}              {rgb}{0.86,0.86,0.86}
\definecolor {gray87}              {rgb}{0.87,0.87,0.87}
\definecolor {gray88}              {rgb}{0.88,0.88,0.88}
\definecolor {gray89}              {rgb}{0.89,0.89,0.89}
\definecolor {gray90}              {rgb}{0.90,0.90,0.90}
\definecolor {gray91}              {rgb}{0.91,0.91,0.91}
\definecolor {gray92}              {rgb}{0.92,0.92,0.92}
\definecolor {gray93}              {rgb}{0.93,0.93,0.93}
\definecolor {gray94}              {rgb}{0.94,0.94,0.94}
\definecolor {gray95}              {rgb}{0.95,0.95,0.95}
\definecolor {gray97}              {rgb}{0.97,0.97,0.97}
\definecolor {gray98}              {rgb}{0.98,0.98,0.98}
\definecolor {gray99}              {rgb}{0.99,0.99,0.99}
\definecolor {darkgrey}            {rgb}{0.66,0.66,0.66}

\newcommand{\GDT}[1]{\textcolor{blue}{\textbf{[GDT: #1]}}}

\renewcommand{\GDT}[1]{\textcolor{black}{#1}}
\newcommand{\changed}[1]{\textcolor{black}{#1}}

\newcommand{\method}{{\texttt{PEAR}}\xspace}
\newcommand{\xmethod}{{\texttt{XPEAR}}\xspace}
\newcommand{\cscf}{{\texttt{CSCF}}\xspace}
\newcommand{\mcts}{{\texttt{MCTS}}\xspace}
\newcommand{\face}{{\texttt{FACE}}\xspace}
\newcommand{\fare}{{\texttt{FARE}}\xspace}
\newcommand{\rl}{{\texttt{RL}}\xspace}
\newcommand{\acronym}{{Preference Elicitation for Algorithmic Recourse}\xspace}

\newcommand{\EU}{\ensuremath{\mathsf{EU}}\xspace}
\newcommand{\EUS}{\ensuremath{\mathsf{EUS}}\xspace}
\newcommand{\FARE}{\ensuremath{\texttt{FARE}}\xspace}
\newcommand{\WFARE}{\ensuremath{\texttt{W-FARE}}\xspace}
\newcommand{\EFARE}{\ensuremath{\texttt{EFARE}}\xspace}
\newcommand{\WEFARE}{\ensuremath{\texttt{W-EFARE}}\xspace}

\title{Personalized Algorithmic Recourse \\ with Preference Elicitation}

\author{\name Giovanni De Toni \email giovanni.detoni@unitn.it \\
      \addr Augmented Intelligence Center, Fondazione Bruno Kessler, Italy \\
      DISI, University of Trento, Italy
      \AND
      \name Paolo Viappiani\email paolo.viappiani@lamsade.dauphine.fr\\
      \addr LAMSADE, CNRS, Université Paris-Dauphine, PSL, France
      \AND
      \name Stefano Teso \email stefano.teso@unitn.it \\
      \addr CIMeC \& DISI, University of Trento, Italy
      \AND
      \name Bruno Lepri \email lepri@fbk.eu \\
      \addr Augmented Intelligence Center, Fondazione Bruno Kessler, Italy
      \AND
      \name Andrea Passerini \email andrea.passerini@unitn.it \\
      \addr DISI, University of Trento, Italy
}

\def\month{01}  %
\def\year{2024} %
\def\openreview{\url{https://openreview.net/forum?id=8sg2I9zXgO}}

\begin{document}
\maketitle

\begin{abstract}
\textit{Algorithmic Recourse} (AR) is the problem of computing a \textit{sequence of actions} that -- once performed by a user -- overturns an undesirable machine decision.
It is paramount that the sequence of actions does not require too much effort for users to implement.  Yet, most approaches to AR assume that actions cost the same for all users, and thus may recommend unfairly expensive recourse plans to certain users.
Prompted by this observation, we introduce \method, the first human-in-the-loop approach capable of providing \textit{personalized} algorithmic recourse tailored to the needs of any end-user.
\method builds on insights from Bayesian Preference Elicitation to iteratively refine an estimate of the costs of actions by asking \textit{choice set} queries to the target user.  The queries themselves are computed by maximizing the \textit{Expected Utility of Selection}, a principled measure of information gain accounting for uncertainty on both the cost estimate and the user's responses.
\method integrates elicitation into a Reinforcement Learning agent coupled with Monte Carlo Tree Search to quickly identify promising recourse plans.
Our empirical evaluation on real-world datasets highlights how \method produces high-quality personalized recourse in only a handful of iterations.
\end{abstract}

\section{Introduction}

Automated decision support systems are increasingly employed in high-risk decision tasks with the aim of empowering human decision-makers and improving the quality of their decisions.
Example applications include bail requests \citep{dressel2018accuracy}, loan approvals \citep{sheikh2020approach}, job applications \citep{liem2018psychology}, and prescription of medications and treatments \citep{yoo2019adopting}.
Despite their promise, often these systems are opaque -- meaning that users, and even engineers, have trouble understanding and controlling their decision process -- and provide no means for overturning unwanted outcomes, such as denied loan requests.
One way of addressing these issues is through the lens of  \textit{Algorithmic Recourse} (AR) \citep{venkatasubramanian2020philosophical, karimi2021algorithmic}.
In AR, given an undesirable machine-generated decision, the goal is to identify a sequence of actions -- or \textit{interventions} for short -- that once implemented by the user overturns said decision, for instance, changing job or obtaining a master's degree.
Motivated by this, a number of approaches have been recently proposed for computing AR \citep{detoni2022synthesizing, naumann2021consequence, karimi2020survey, ramakrishnan2020synthesizing, shavit2019extracting, ustun2019actionable,
karimi2020algorithmic, tsirtsis2021decisions, russell2019diverse, mothilal2020explaining, wang2023gam, dandl2020multi}.

It is critical that the suggested recourse plans are not too difficult or expensive to carry out.  This entails that recourse should be \textit{personalized}, because different users in the same situation may \textit{need} substantially different recourse plans.
To see this, consider a user who is denied a loan.  Based on a profile made by the financial institution, an AR algorithm might suggest the user reduce their monthly expenses.  However, unlike the ``average'' customer, our user is incurring high medical expenses because they recently contracted an invalidating illness.  Thus, the AR suggestion is highly inappropriate.  Clearly, it is impossible to infer such a constraint from their profile alone.
Most approaches, however, completely neglect the user's own preferences.  The few that do require feedback that is difficult to obtain in practice, \eg preferences over a large pool of alternatives \citep{russell2019diverse, mothilal2020explaining, wang2023gam, dandl2020multi} or upfront quantification of action costs~\citep{yadav2021low, karimi2020survey, rawal2020beyond, wang2023gam, mahajan2019preserving}.

\begin{figure}[!t]
    \centering
    \includegraphics[width=0.7\linewidth]{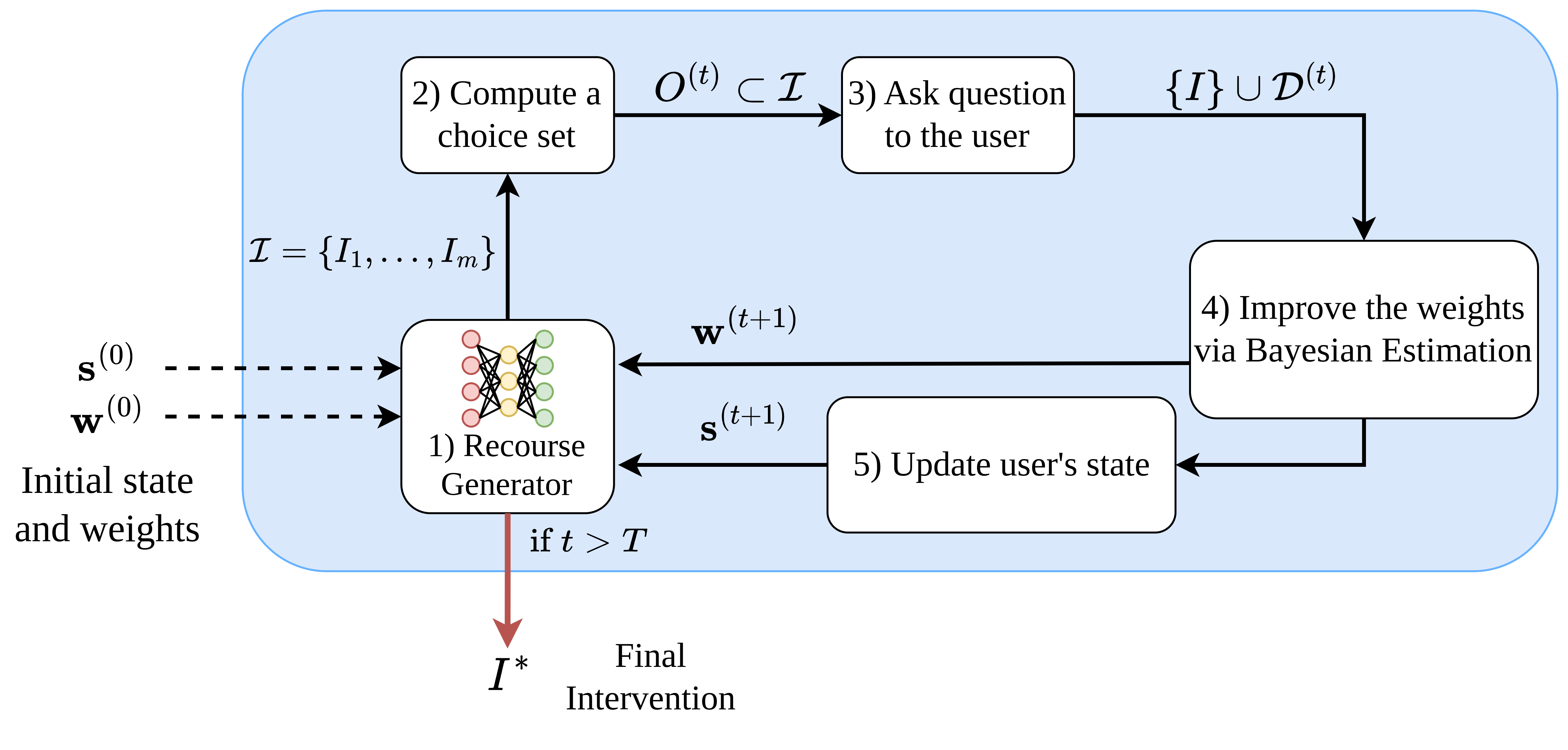}
    \caption{Overview of \method.
    (1) Given the initial state $\vs^{(0)}$  of the user and weights $\vw^{(0)}$, \method computes a pool of candidate interventions achieving recourse.
    (2) A choice set $O^{(t)}$ is selected from the pool and presented to the user.
    (3) The user picks their preferred intervention from the set.
    (4) An improved estimate of the weights $\vw^{(t+1)}$ is computed using this feedback, and
    (5) the user's state $\vs^{(t+1)}$ is updated.
    After $T$ rounds, the estimated weights are used to compute a final intervention $I^{*}$.
    }
    \label{fig:procedure_complete}
\end{figure}

We argue that algorithmic recourse should make users first-class citizens in the recourse generation process rather than viewing them as passive observers.
To this end, we introduce \method (\acronym), the first human-in-the-loop approach for generating \textit{personalized} recourse tailored for a target end-user.
Our algorithm integrates AR and ideas from interactive Preference Elicitation (PE) \citep{chajewska2000making, boutilier2002pomdp, braziunas2007minimax, guo2010real, viappiani2010optimal, dragone2018constructive} in a fully Bayesian setup.
\method goes beyond existing approaches in that the costs of actions are estimated from user feedback and prior information.  In each iteration, \method identifies a \textit{small} selection of alternative interventions -- a \textit{choice set} -- that optimizes a sound measure of information gain (the \textit{Expected Utility of Selection} (\EUS) \citep{viappiani2010optimal,viappiani2020equivalence}) and then asks the user to pick their preferred option.  Using this feedback, \method quickly improves its estimate of the user's preferences and generates interventions that get progressively closer to the user's ideal. Furthermore, \method takes interactions and dependencies between actions costs into account when computing suggested recourse, whenever this information is available.
See \cref{fig:procedure_complete} for an overview.

\textbf{Contributions}:  Summarizing, we:
\begin{itemize}[leftmargin=1.25em]

    \item Introduce the problem of \textit{personalized algorithmic recourse}, and show that existing approaches are insufficient to solve it.

    \item Develop \method, the first human-in-the-loop, Bayesian approach for computing \textit{personalized} interventions that is robust to noise in user feedback and minimizes user effort.

    \item Evaluate \method on synthetic and real-world datasets and show that it can generate substantially -- up to 50\% -- cheaper interventions than user-agnostic competitors after only a handful of queries.

\end{itemize}

\section{Problem Statement}
\label{sec:problem-statement}

The user \textit{state} $\vs = (s_1, \ldots, s_d) \in \calS \subseteq \bbR^d$ is a vector of $d$ categorical and real-valued features encoding, \eg instruction level and income.
An \textit{action} $a \in \calA$ is a map that takes a state $\vs$ and changes a \textit{single} feature, yielding a new state $\vs' = a(\vs)$, and expresses a recommendation of the form ``\textit{Increase your income by \$100}.''
Given a (black-box) binary classifier\footnote{
It is straightforward to adapt our approach to deal with multiclass classification problems.} $h: \calS \rightarrow \{0, 1\}$ and a user state $\vs$ leading to an undesirable decision $y = h(\vs)$, AR computes an \textit{intervention} -- \ie an ordered set of actions $I = \{a^{(1)}, \ldots, a^{(|I|)}\}$ -- that can be applied to $\vs$ to obtain a counterfactual state $\vs'$ associated to a more desirable outcome $h(\vs') \ne y$, \textit{all while minimizing user effort}.
\GDT{We formalize the user effort required to perform an action via a \textit{cost function} $C: \calA \times \calS \to \bbR^+$. The cost of an intervention is the sum of the costs of all actions it contains, that is $C(I) = \sum_{i=0}^{|I|} C(a^{(i)}, \vs^{(i)})$. Here, $\vs^{(i)}$ is the state obtained by applying action $a^{(i-1)}$ to state $\vs^{(i-1)}$, and $\vs^{(0)} = \vs$ is the initial state. We denote with $I(\vs^{(0)})$ the operation of applying each action $a \in I$ sequentially.}

Approaches to AR assume the user effort is proportional to the number of actions that need to be carried out, and minimize it by searching for \textit{short} interventions $I$.
However, this is unrealistic and impractical. For instance, changing job into a highly skilled one may not be realistic without obtaining a Master’s degree first.
\GDT{In reality, the user effort also depends on \textit{users' preferences} $\vw \in \calW \subseteq \bbR^d$ which we represent as $d$-dimensional vectors. For example, for some people, it might be easier to improve their education than get new employment.  Thus, we \textit{parameterize} the cost function using the user preferences as $C(I \mid \vw)$, or $C(a, \vs \mid \vw)$ if we refer to single actions.
In the following, we distinguish between the (typically unknown) user preferences $\vw^{GT}$ and the preferences $\vw$ used by the recourse algorithm, which might differ from the former.
}
Motivated by this, we introduce a new problem setting, denoted \textit{personalized algorithmic recourse}:

\begin{definition}[Personalized Algorithmic Recourse]
\label{def:par}
Given a black-box binary classifier $h$ and a user state $\vs$, acquire a cost function $C(I \mid \vw)$ for interventions such that the intervention $I^*$ obtained by solving the following optimization problem:
\begin{equation}
    \textstyle
    I^* \in \argmin_{I} \; C(I \mid \vw)
    \quad \mathrm{s.t.} \quad
    h(I(\vs^{(0)})) \neq h(\vs^{(0)})
    \label{eqn:par}
\end{equation}
has \textbf{minimal regret} for the target user, defined as:
\[
    \textstyle
    Reg(I^*,I^\textrm{GT}) = C(I^* \mid \vw^\textrm{GT}) - C(I^\textrm{GT} \mid \vw^\textrm{GT})
    \label{eqn:par_regret}
\]
where $\vw^\textrm{GT} \in \calW$ encodes the ground-truth but unobservable preferences of the user and $I^\textrm{GT}$ is the ``ideal'' intervention that would be obtained by solving \cref{eqn:par} using $\vw^\textrm{GT}$.
\end{definition}

The similarity of \cref{eqn:par} to existing formulations of AR can be misleading, as here the key challenge is that of obtaining weights $\vw$ that reflect the user's own preferences.
We discuss how \method does so in \cref{sec:method}.

\section{Related work}
\label{sec:related-work}

\textit{Counterfactual explanations} (CEs) are a class of local, human-understandable explanations \citep{wachter2017counterfactual, byrne2019counterfactuals} that convey information about changes to input variables that overturn a machine decision \citep{guidotti2018lore, stepin2021survey}.  AR aims to identify \textit{actionable} CEs that attain recourse for the user \citep{verma2020counterfactual, karimi2020survey}.
Existing approaches to AR solve \cref{eqn:par} via diverse optimization methods \citep{wachter2017counterfactual, ramakrishnan2020synthesizing, poyiadzi2020face, naumann2021consequence, russell2019diverse, mothilal2020explaining, wang2023gam, dandl2020multi} or by learning a general policy \citep{shavit2019extracting, detoni2022synthesizing, verma2022amortized}.
Most methods simply return a set of actions, disregarding their order. However, recent research showed that ignoring the causal relationship between features prevents reaching optimal recourse \citep{karimi2021algorithmic}. Some methods thus optimize for recourse plans, \ie \textit{sequences} of actions attaining recourse, following a purely causal or causal-inspired setup \citep{ustun2019actionable, karimi2021algorithmic, karimi2020algorithmic, detoni2022synthesizing, naumann2021consequence, mahajan2019preserving}.
\method fully supports this paradigm and considers the interplay between features when finding recourse, whenever this information is available.

Most AR approaches assume that the cost function is fully specified beforehand \citep{wachter2017counterfactual, ramakrishnan2020synthesizing, rawal2020beyond, wang2023gam, mahajan2019preserving}, ignoring the problem of modelling user preferences altogether. The few that explicitly deal with user preferences do so in a na\"ive manner.
Some of them \citep{russell2019diverse, mothilal2020explaining, dandl2020multi, yadav2021low} ask users to pick their preferred option from a \textit{large} pool of user-agnostic recourse plans, that is not guaranteed to contain a low-cost option for the user.  This is also impractical, as users can only properly evaluate a limited number of alternatives at a time \citep{simon1955behavioral, march1978bounded}.
Others require users to \textit{quantify} the cost of each possible action upfront \citep{karimi2020survey, rawal2020beyond, wang2023gam, mahajan2019preserving}, or via numerical constraints \citep{poyiadzi2020face, wang2023gam}, yet end-users can rarely articulate their preferences in a quantitative manner \citep{Keeney1993}.
\method sidesteps this issue by learning preferences from ranking data, \ie relative judgments of the form ``I prefer option $A$ to option $B$''~\citep{FurHul2010}.

AR is specifically concerned with high-stakes scenarios, such as loan requests, that users face at most a handful of times in their lifetime.  In these settings, it is impossible to estimate user preferences from historical data.
In recommender systems, this issue has been solved through \textit{preference elicitation}, whereby the user's preferences are estimated through a human-friendly interaction protocol \citep{boutilier2002pomdp}.
In order to converge to high-quality options with minimal user effort, PE algorithms select queries (\ie questions to the user) that maximize information gain and that are easy to answer.
A popular option is \textit{choice queries}, in which the user has to select a preferred item from a small set of alternatives.
\method builds on Bayesian PE methods, as they account for imprecision in users' answers \citep{chajewska2000making, guo2010real, viappiani2010optimal} in a principled way by measuring the information gain of choice sets in terms of Expected Utility of Selection \citep{viappiani2020equivalence}. %
Similarly to \method, \citet{rawal2020beyond} estimate cost weights from preference feedback. However, they collect feedback from \textit{domain experts} in a non-interactive fashion and learn \textit{population-level} preferences that are not personalized, thus failing to minimize \cref{eqn:par_regret}. Population-level estimates can lead to recourse that is largely suboptimal for specific individuals, as will be shown in our experimental evaluation.

\section{Personalized Algorithmic Recourse with \method}
\label{sec:method}

\begin{figure}[t]
    \centering
    \begin{subfigure}[t]{0.4\textwidth}
        \includegraphics[width=\textwidth]{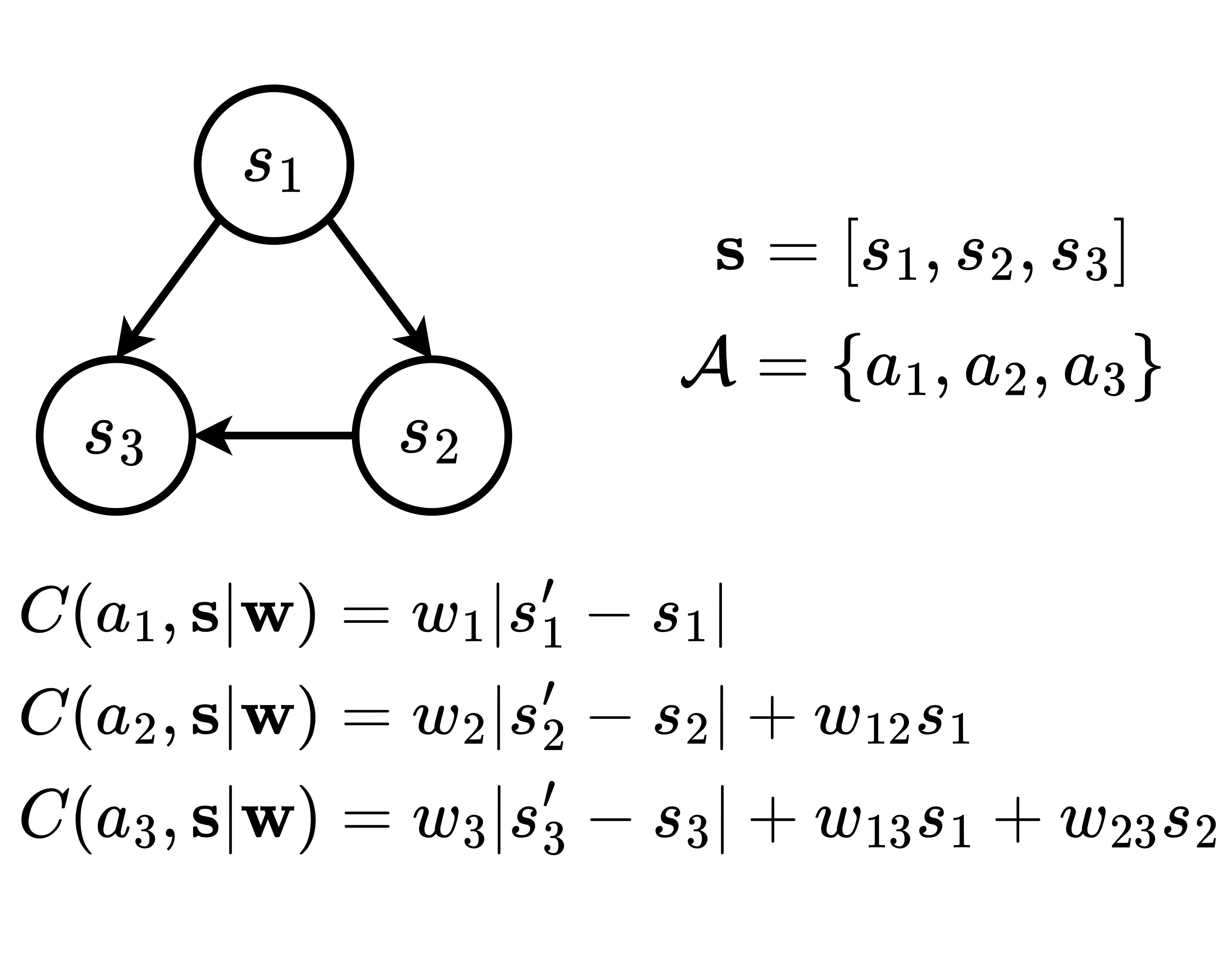}
        \caption{Cost Correlation Structure (CCS)}
        \label{fig:cost-correlation-structure}
    \end{subfigure}%
    \hfill
    \begin{subfigure}[t]{0.55\textwidth}
        \includegraphics[width=\textwidth]{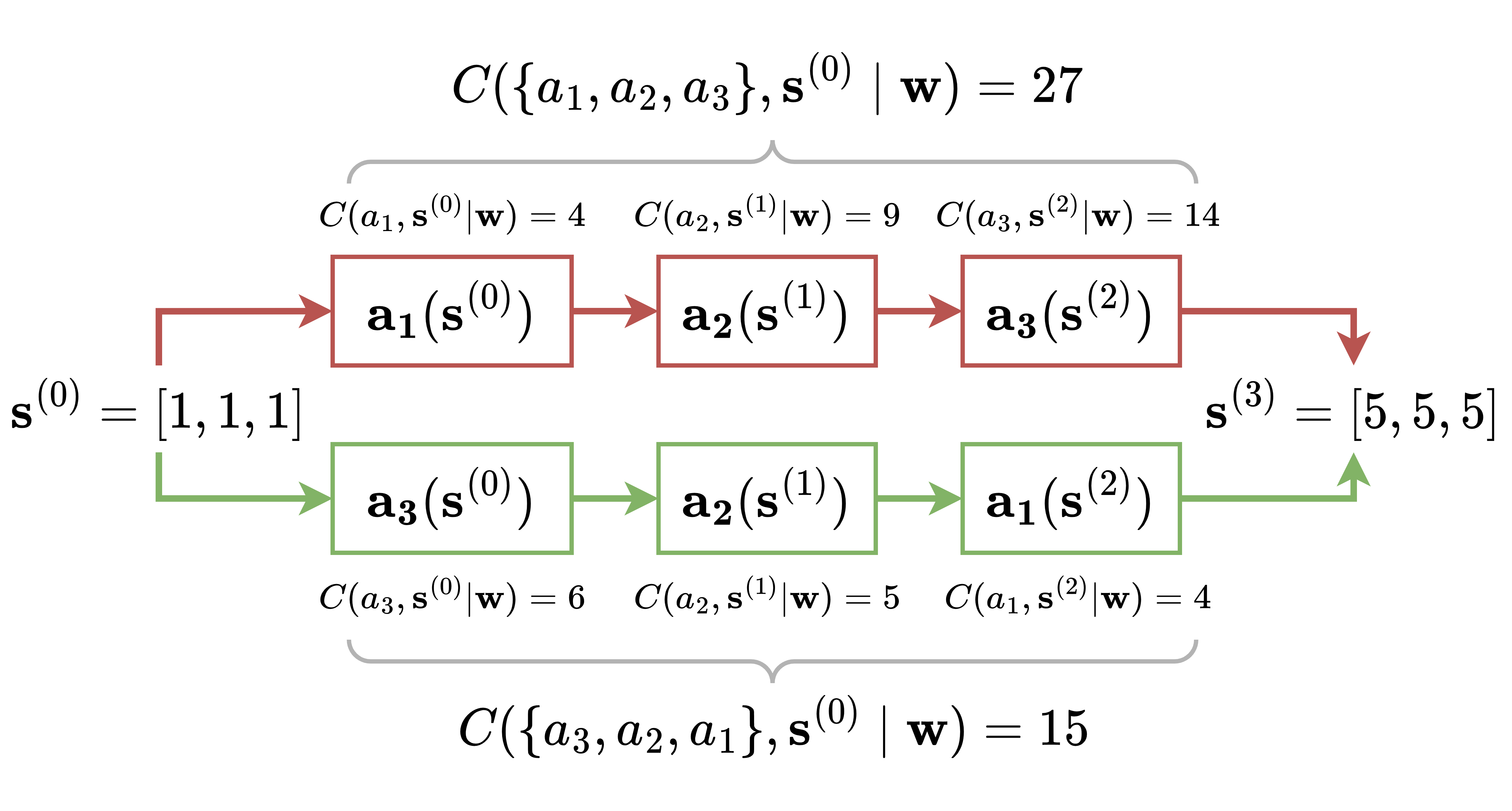}
        \caption{The cost of an intervention $I$}
        \label{fig:intervention-ordering}
    \end{subfigure}
    \caption{\textbf{The cost model} (Left) A \textit{cost correlation structure}
 (CCS) for cost modelling.
    (Right) Given $\vs^{(0)} = [1,1,1]$ and unit $\vw$, let us imagine we want to reach $\vs^{(3)} = [5,5,5]$ by following the presented CCS and the intervention $I=\{a_1, a_2, a_3\}$, where $a_i$ assign $s_i \leftarrow 5$ for all $i \in \{1,2,3\}$. Clearly, we incur different costs by applying permuted versions of $I$. The green path indicates the lower-cost intervention.}
    \label{fig:scm-cost}
\end{figure}

Before describing \method, we discuss how we model the user's preferences and how these impact the costs of actions and interventions.
Equally importantly, actions influence each other's costs -- \eg obtaining an additional degree can dramatically lower the cost of landing a better-paying job -- and we need to account for this if we wish to minimize the cost of recourse.
In order to account for interactions between action costs, we model $C$ using a set of linear equations, parameterized by weights $\vw \in \calW$, \GDT{by taking inspiration from \textit{generalised additive independence} models \citep{pigozzi2016preferences} from decision theory and \textit{structural causal models} \citep{pearl2009causality}. We call such formalization \textit{cost correlation structure} (CCS).}
A CCS corresponds to a directed acyclic graph (DAG) where each node is associated with the cost $w_k$ of changing a single feature $s_k$ and each edge represents a (causal) relationship between two costs, parameterized by $w_{jk} \in \bbR$.
Then, following the equation induced by the DAG, we define the cost of applying an action $a_k$ to a state $\vs$ to change the $k$-th feature from $s_k$ to $s_k'$ as a linear combination:
\[
    \textstyle
    C(a_k, \vs \mid \vw)
        = w_k |s_k'-s_k| + \sum_{j \; \in\; Pa_k} w_{jk} s_j
    \label{eqn:cost}
\]
Here, $Pa_k$ are the \textit{parents} of the $k$-th node in the graph.
\GDT{For example, we can use \cref{eqn:cost} to describe the cost of the action "\textit{increase your salary}". Such cost depends linearly on the raise asked ($|s_k'-s_k|$). However, it also depends on some causally related variables ($\sum_{j \; \in\; Pa_k} w_{jk} s_j$), such as the user's current job. We can imagine that it might be easier to ask for a raise for a company's manager rather than a simple employee.}
A formal example of a cost function and of its corresponding DAG is shown in \cref{fig:cost-correlation-structure}.
Again, the cost of an intervention is the sum of the costs of all actions it contains applied sequentially, that is:
\[
    \textstyle
    C(I \mid \vw) = \sum_{i=0}^{|I|} C(a^{(i)}, \vs^{(i)} \mid \vw)
    \label{eqn:intervention_cost}
\]
\cref{eqn:cost} does not define a simple linear model, but it accounts for a richer set of interactions. For example, given two actions $a_1$ and $a_2$, their cost is not additive, meaning $C(a_1, \vs \mid \vw) + C(a_2, \vs \mid \vw) \neq C(\{ a_1, a_2 \}, \vs \mid \vw)$.
We can show it with a simple example.

\textbf{Example of non-additivity of \cref{eqn:cost}.} Consider a simple CCS with two nodes $s_1 \rightarrow s_2$, with $\vs = (s_1, s_2)$ and $\vw = (w_1, w_2, w_{12})$, and an instance $\vs = (1, 1)$ and $\vw = (1, 0.5, 1)$.  We define two actions, $a_1$ and $a_2$, which simply set the corresponding features $s_i$ to 2. Using \cref{eqn:cost}, we can compute the cost of applying the single action as
\[
C(a_1, \vs \mid \vw) = 1|2-1| = 1 \qquad C(a_2, \vs \mid \vw) = 0.5|2-1|+1 = 1.5
\]
By \cref{eqn:intervention_cost}, the cost of the intervention $\{a_1, a_2\}$ is $C(\{a_1, a_2\}, \vs \mid \vw) = 3.5 \ne 1.5 + 1$, hence the costs are \textit{not} additively independent.
Moreover, \cref{eqn:intervention_cost} considers the \textit{order} in which actions are applied, \ie $\exists \; \vw$ such that $C(\{a_1, a_2\}, \vs \mid \vw) \neq C(\{a_2, a_1\}, \vs \mid \vw)$, as shown in \cref{fig:intervention-ordering}.

Note that if dependencies between actions costs are not available, \cref{eqn:cost} boils down to a simple linear function, namely $C(a_k, \vs \mid \vw) = w_k | s'_k - s_k |$.  This case matches the typical setup of algorithmic recourse methods, which assume the weights are independent from one another.  \cref{eqn:cost} is however more flexible in that, \textit{if} dependencies between costs \textit{are} in fact available, they are automatically leveraged by \method.

\color{black}

We formalized the cost of an action (\cref{eqn:cost}) by following a standard assumption made in decision theory \citep{Keeney1993} called "\textit{generalised additive independence}" (GAI) \citep{pigozzi2016preferences} which is shared by many preference elicitation algorithms. It defines utility functions that measure both the contribution of a single feature and the contribution of a subset of features. Following causality theory \citep{pearl2009causality}, each subset consists of features causally related to the one we are acting on. In support of this strategy, previous research highlighted how it is impossible to offer optimal recourse recommendations without considering relationships between features \citep{karimi2020algorithmic}.

In \citet{karimi2020algorithmic}, the authors refer to knowing the impact of an intervention on causally related features (e.g., will getting a new degree increase my salary?). However, it is known that such causal knowledge about the world is hard to obtain unless we make specific restricting assumptions \citep{pearl2009causality}. In our work, rather than considering how features causally change after an intervention, we focus on learning how the "cost" of acting on those features changes after an intervention (e.g., will getting a new degree make it easier to increase my salary?). In our case, the cost of modifying a feature is something we can easily learn by asking the user.

Ultimately, with our setup, we make learning the cost function feasible, and we can also represent complex non-additive behaviour which arises from the sequential nature of recourse.
Our formalization in \cref{eqn:cost} and \cref{eqn:intervention_cost} goes along the lines of previous works \citep{naumann2021consequence, detoni2022synthesizing}, that address the problem from the same perspective, but assume to know these costs a-priori.

\color{black}

\textbf{On Cost Correlation Structure and Causality.} The \textit{cost correlation structure} shares some similarities with Structural Causal Models (SCMs) \citep{pearl2009causality}, since they both use a directed acyclic graph, also called \textit{causal graph}, to represent causal relationships between features $\vX$.
In the context of AR, SCMs model relationships between \textit{features} rather than between \textit{costs}. For example, \citet{karimi2020algorithmic} frames the recourse problem as finding a counterfactual $\vs^{\bbC\bbF}$ optimizing \cref{eqn:par} given an approximated SCM describing the joint distribution $P(\vX)$, while the cost function is a simple $p$-norm, such as $C(\vs^{\bbC\bbF}, \vs) = \| \vs^{\bbC\bbF} - \vs \|_2$.
In our case, CCSs do not allow for $do$-actions or to compute counterfactuals via abduction, since they do not model $P(\vX)$.
Similarly to \citet{naumann2021consequence}, we use the causal DAG to represent consequence-aware cost equations as shown in \cref{eqn:cost}, without taking any explicit causal angle. Indeed, following the non-causal AR literature, we assume actions $a \in \calA$ to influence only the target feature $s_k$, without any causal effect on any of its children $s_j$ in the DAG. However, actions over $s_k$ will affect the \textit{cost} of subsequent actions changing the user features.

\subsection{The \method Algorithm}

In order to account for uncertainty over the user's weights $\vw$, \method explicitly models a distribution $P(\vw)$ over them and progressively refines it by interacting with a target user.  A high-level overview of \method is given in \cref{fig:procedure_complete} and the pseudo-code is listed in \cref{alg:complete_procedure}.

In each iteration $t = 1, \ldots, T$, where $T$ is the iteration budget, \method computes a \textit{choice set} $O^{(t)} \in \calI^k$ containing $k$ candidate interventions achieving recourse (for a small $k$, \eg $2$ to $4$) and asks the user to indicate their most preferred option in the set.
Importantly, $O^{(t)}$ is chosen so as to maximize the (expected) information gained from the user, and in a way that is robust to noise in their feedback.  We detail the exact procedure used by \method in \cref{sec:computing-choice-sets}.
These user choices are stored in an initially empty dataset $\calD^{(t)}$.  In each step, \method integrates the user's feedback by inferring a posterior over the weights $P(\vw \mid \calD^{(t)}) \propto P(\calD^{(t)} \mid \vw)P(\vw)$ using Bayesian inference, %
and updates the user state by applying the first action $\hat{I}_{1}$ of the chosen intervention. We apply a single action so as to elicit user preferences in all intermediate states. If a state achieving recourse is reached, the user state is reinitialized.
After $T$ rounds,\footnote{In practice, the loop can be terminated as soon as the user is satisfied with one of the interventions in $O^{(t)}$.} \method computes a low-cost personalized intervention by applying the intervention generation procedure described in \cref{sec:computing-interventions}, biased according to the latest posterior $p(\vw \mid \calD^{(t)})$.

\method makes no assumption on the form of the prior $P(\vw)$, meaning that the prior can be adjusted based on the application.
In order to model both variances across the preferences of individuals and for sub-groups in the population, in this work we model them as a mixture of Gaussians with $M$ components $\mathcal{N}_i(\mathbbm{\mu}_i,\mathbf{\Sigma}_i)$, for $i = 1, \ldots, M$.  This choice works well in our experiments, see \cref{sec:experiments}.
Note that, analogously to \citet{rawal2020beyond}, it is also possible to fit the prior on population-level preference data or domain expert input, whenever this is available.  In our experiments, we do this for \textit{all} competitors.

\subsection{Generating Personalized Interventions with \WFARE}
\label{sec:computing-interventions}

\method generates personalized interventions by leveraging a novel, \textit{user-aware} extension of \FARE \citep{detoni2022synthesizing}, a state-of-the-art algorithm for generating short -- but \textit{user-agnostic} -- interventions, which we briefly outline next.
In \FARE, each action $a \in \calA$ is implemented as a tuple $(f, x)$, where $f$ is a function changing \textit{one} feature and $x$ is the value that feature takes, \eg $(change\_income, \$1000)$.
Given an initial state $\vs$, \FARE uses reinforcement learning to learn two probabilistic policies $\pi_{f}(\vs)$ and $\pi_{x}(\vs)$, which are used as priors to guide a Monte Carlo Tree Search procedure that incrementally builds an intervention $I$ by selecting actions $a^{(i)} \in \calA$.  In order to ensure interventions are \textit{actionable}, actions $a$ are only chosen if they satisfy given preconditions.
The reward used by \FARE is $r(I) = \rho^{|I|} \cdot \Ind{h(I(\vs^{(0)})) \neq h(\vs^{(0)})}$, where $\rho > 0$ is a discount factor and the indicator evaluates to $1$ if $I$ attains recourse and to $0$ otherwise.
\FARE is highly scalable and very effective at identifying counterfactual interventions even under minimal training budget \citep{detoni2022synthesizing}.

\FARE is user-agnostic, while \method needs to generate \textit{personalized} interventions.
We fill this gap by introducing \WFARE, a novel extension of \FARE that integrates the user's costs into the reward while inheriting all benefits of the latter.
Recall that \method maintains a posterior over the weights.  The \textit{expected cost of an action} can thus be obtained by marginalizing over the posterior:
\[
    \textstyle
    \bbE[C(a, \vs) \mid \calD^{(t)}]
    = \int_{\vw} C(a, \vs \mid \vw) P(\vw \mid \calD^{(t)}) \; d\vw
    \label{eqn:exp_action_cost}
\]
Analogously, the cost of an intervention $I$ is replaced by the expectation:
\[
    \textstyle
    \bbE[C(I) \mid \calD^{(t)}] = \sum_{i = 0}^{|I|} \bbE[C(a^{(i)}, \vs^{(i)}) \mid \calD^{(t)}]
    \label{eqn:exp_intervention_cost}
\]
The \WFARE reward function is then given by $r(I \mid \vw) \propto \rho^{\bbE[C(I \mid \calD^{(t)})]} \cdot \Ind{h(I(\vs^{(0)})) \neq h(\vs^{(0)})}$ (see \cref{sec:wfare-full} for a more in-depth description of \WFARE).
This explicitly drives RL to learn policies that optimize user-specific action costs and that, therefore, help MCTS to more quickly converge to \textit{personalized} interventions.
We show empirically that \method is substantially more effective than \FARE at computing personalized interventions in \cref{sec:experiments}.

\begin{algorithm}[!t]
    \caption{The \method algorithm:  $h: \calS \rightarrow \{0,1\}$ is a classifier, $\vs^{(0)} \in \calS$ the initial state, $\calA$ the available actions, $p(\vw)$ the prior, $T \ge 1$ the query budget, $k \ge 2$ is the size of choice sets.}
    \label{alg:complete_procedure}
    \begin{algorithmic}[1]
        \Procedure{\method}{$h, \vs^{(0)}, \calA, T, k$}
            \State Initialize $t \gets 0, \; \calD^{(0)} \gets \emptyset$
            \For{$t = 1, \ldots, T$}
                \State $O^{(t)} \gets \text{\tt SUBMOD-CHOICE}(h, \vs^{(t-1)}, \calA, k, \calD^{(t-1)})$ \Comment{\cref{alg:choice_set_greedy}}
                \State Ask the user to pick the best intervention $\hat{I} \in O^{(t)}$
                \State $\calD^{(t)} \gets \calD^{(t-1)} \cup \{ \hat{I} \}$
                \State Update weight estimate $p(\vw \mid \calD^{(t)})$
                \State $\vs^{(t)} \leftarrow \hat{I}_{1}(\vs^{(t-1)})$
                \If{$h(\vs^{(t)}) \ne h(\vs^{(0)})$}
                    \State $\vs^{(t)} \gets \vs^{(0)}$
                \EndIf
            \EndFor
            \State $I^* = \WFARE(h, \vs^{(0)}, \vw^*)$ with $\vw^* = \bbE_{P(\vw \mid \calD^{(T)})}[\vw]$
            \Comment{\cref{sec:computing-interventions}}
            \State \Return $I^{*}$
        \EndProcedure
    \end{algorithmic}
\end{algorithm}

\subsection{Computing Informative Choice Sets}
\label{sec:computing-choice-sets}

Given the current posterior $P(\vw \mid \calD^{(t)})$, \method computes a \textit{choice set} containing $k$ interventions $I$ that maximizes information gain \citep{chajewska2000making}.
We measure the latter using the \textit{Expected Utility of Selection} (\EUS) \citep{viappiani2020equivalence}, a measure of the goodness of a set defined as the expectation, under the uncertainty over $\vw$, of the utility of its most preferred element.
EUS is closely related to the Expected Value of Information (EVOI), and frequently used in Bayesian PE \citep{Price2005, viappiani2010optimal, Akrour2012, viappiani2020equivalence}.
The \EUS builds on the notion of \textit{expected utility of an intervention} $I$, which is defined as:
\[
    \textstyle
    \EU(I \mid \calD^{(t)})
        = \bbE[ - C(I) \mid \calD^{(t)}]
        = - \int_{\vw} C(I \mid \vw) P(\vw \mid \calD^{(t)}) \; d\vw
    \label{eq:eu}
\]
The $\EUS$ of a choice set $O$ can then be defined as:
\begin{align}
    \EUS_R(O \mid \calD^{(t)})
        & \textstyle
            = \sum_{I \in O}
            P_R(O \leadsto I) \EU(I \mid \calD^{(t)})
            \nonumber
    \\
        & \textstyle
            = - \int_{\vw} \left[
                \sum_{I \in O}
                    P_R(O \leadsto I \mid \vw) C(I \mid \vw)
                \right]
                P(\vw \mid \calD^{(t)})
                \; d \vw
            \label{eq:eus}
\end{align}
Here, $P_R(O \leadsto I \mid \vw)$ is the probability that a user with weights $\vw$ picks $I$ from $O$, under a specific choice of \textit{response model} $R$ modelling noise in user choices.
Intuitively, we expect users to prefer the cheapest interventions $I \in O$. Moreover, we also expect that interventions in $O$ with similar costs have a similar probability of being chosen.
Motivated by this, and following common practice in choice modelling \citep{luce2012individual}, in \method we implement a \textit{logistic response model} (\textit{L}), defined as:
\[
    P_{L}(O \leadsto I \mid  \vw)
    =
    \frac{\exp(-\lambda C(I \mid \vw))}{\sum_{I \in O} \exp(-\lambda C(I \mid \vw))}
    \label{eq:logistic}
\]
Here, $\lambda \in \bbR$ is a temperature parameter.
Finding a choice set $O$ maximizing the $\EUS$ is intractable in general -- \textit{NP-hard} \citep{nemhauser1978analysis, krause2014submodular}, in fact -- and computationally intensive in practice, and risks slowing down the interaction loop to the point of estranging users.
We observe that, however, under some response models $R$, the $\EUS$ becomes \textit{submodular} and \textit{monotonic} \citep{viappiani2010optimal}.
This is the case for the \textit{noiseless} response model (\textit{NL}), according to which the user always prefers the lowest-cost option, \ie\footnote{For the sake of presentation, we assume that there are no ties. Note that the \EUS formula is invariant to the way ties are broken. In our implementation, ties are broken uniformly at random.}
\[
    \textstyle
    P_{NL}(O \leadsto I \mid \vw)
    =
    \prod_{I, I' \in O \; : \; I \ne I'} \Ind{ C(I \mid \vw) < C(I' \mid \vw) }
    \label{eqn:noiseless}
\]
This means that, for \textit{NL}, greedy optimization is sufficient to find a choice set $O$ that achieves high $\EUS_{NL}$ with approximation guarantees.  Formally, it holds that for choice sets $O$ found via greedy optimization, $\EUS_{NL}(O \mid  \calD) \geq (1 - e^{-1})\EUS_{NL}(O^* \mid  \calD)$, where $O^*$ is the truly optimal choice set \citep{viappiani2010optimal, nemhauser1978analysis, krause2014submodular}.
In \method, we leverage the fact that $\EUS_{L} - \EUS_{NL}$ is always smaller than a problem independent (tight) bound \citep{viappiani2010optimal}, meaning that instead of minimizing $\EUS_{L}$ directly, we can compute a high-quality choice set by greedily maximizing $\EUS_{NL}$.
This immediately leads to a practical algorithm for the logistic response model $L$, listed in \cref{alg:choice_set_greedy}.

\begin{algorithm}[!t]
    \caption{Greedy procedure to efficiently compute a choice set $O$: $\vs^{(t)} \in \calS$ the current state, $\calA$ the available actions, $k \ge 2$ is the size of choice sets, $\calD^{(t)}$ the user choices so far.}
    \label{alg:choice_set_greedy}
    \begin{algorithmic}[1]
        \Procedure{\texttt{SUBMOD-CHOICE}}{$\vs^{(t)}, k, \calA, \calD^{(t)}$}
            \State $O \gets \emptyset$
            \State $\bar\vw \gets \bbE_{p(\vw \mid \calD^{(t)})}[\vw]$
            \While{$|O| < k$}
                \State Generate the candidate interventions $\calI$ with \WFARE using $\calA$ and $\bar\vw$
                \State $\hat{I} \gets \hbox{argmax}_{\hat{I}} \;\; \EUS_{NL}(O \cup \hat{I} \mid  \calD^{(t)}) - \EUS_{NL}(O \mid \calD^{(t)})$
                \State $O \gets  O \cup \{\hat{I}\}$
            \EndWhile
            \State \Return $O$
        \EndProcedure
    \end{algorithmic}
\end{algorithm}

\subsection{Benefits and limitations}  \method is designed to facilitate the application of AR to practical high-stakes tasks like loan approval.
The main benefit of \method is that it provides \textit{personalized} algorithmic recourse, which existing approaches are not capable of.  Also, it follows a fully Bayesian setup for handling uncertainty over the estimated preferences and noise in user feedback.
It also leverages ideas from preference elicitation -- such as small \textit{choice sets} and elicitation of relative preferences -- to ensure the interaction is cognitively affordable.
Finally, \method makes use of a novel cost function that automatically takes dependencies between action costs into consideration, whenever these are known or can be estimated, thus facilitating the identification of better recourse suggestions.  At the same time, the cost function is linear, enabling \method to leverage efficient algorithms from preference elicitation tailored for eliciting such cost functions.

Several steps of the algorithm -- namely, evaluating the \EUS (\cref{eq:eus}) and the expected cost of interventions (\cref{eqn:exp_intervention_cost}), and updating the posterior $p(\vw \mid \calD^{(t)})$ -- require marginalizing over the weights.
Doing so involves evaluating a complex integral that cannot be solved analytically.
We sidestep this issue by leveraging an efficient Monte Carlo approximation, and specifically \textit{ensemble split sampling} \citep{karamanis2020ensemble, karamanis2021zeus}  and then averaging over the samples with the highest likelihood.
We find that, empirically, this procedure is efficient -- so much that it can support interactive usage -- and leads to competitive results in our experiments, cf. \cref{sec:experiments}.

\GDT{
Lastly, as is typical in PE, our approach focuses on the “myopic” value of information by looking to maximize the expected posterior utility, thus reducing as much as possible the regret \textit{one step ahead}. As we ask more questions, the regret will tend to zero. However, one main theoretical challenge currently faced in the PE domain is the development of optimal elicitation policies that consider a \textit{sequential} version of the value of information, which may be more efficient but intractable in general \citep{boutilier2002pomdp}.
}

\begin{figure}
    \centering
    \includegraphics[width=\textwidth]{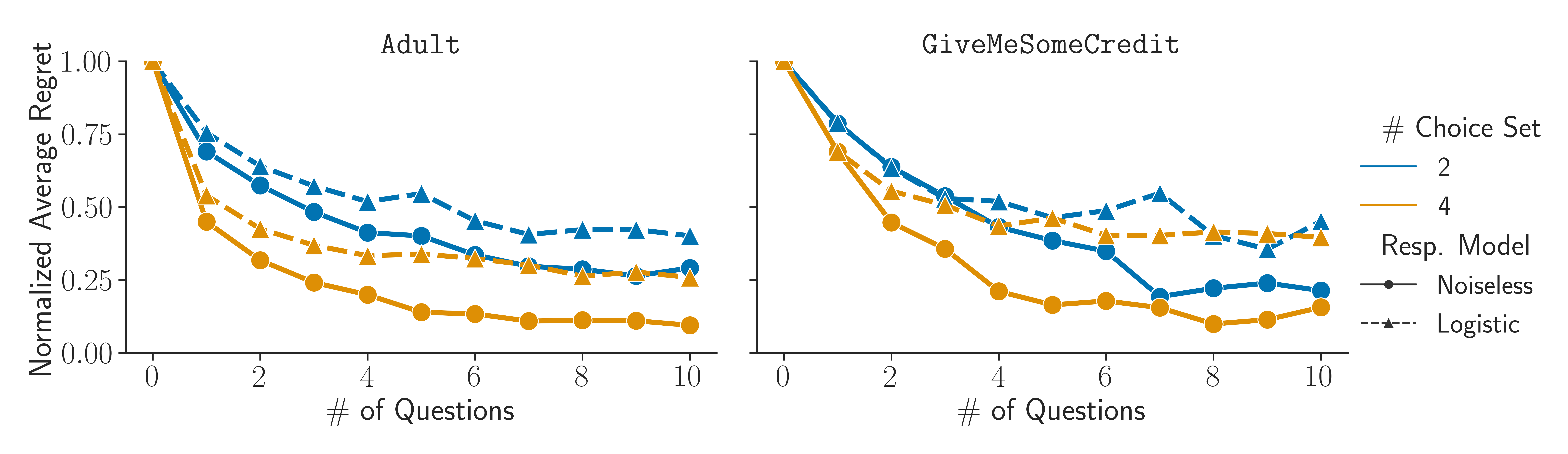}
     \caption{Normalized Average Regret for \method when varying the number of questions, the choice set size and the user response model on both datasets (sampled from \texttt{All} users).} %
     \label{fig:average-regret}
\end{figure}

\section{Experiments}
\label{sec:experiments}

Our experimental evaluation is aimed at answering the following research questions:

\begin{itemize}
    \item[{\bf Q1}] Does \method succeed in minimizing regret for an increasing amount of user feedback?
    \item[{\bf Q2}] Does \method outperform competitors in terms of validity and cost?
    \item[{\bf Q3}] Is \method robust to imprecise knowledge of the cost correlation structure of the user?
\end{itemize}

\textbf{Datasets and Classifiers.} We evaluated our approach on two real-world datasets taken from the relevant literature: \texttt{GiveMeSomeCredit} \citep{GiveMeSomeCredit} and \texttt{Adult} \citep{Dua:2019}.
They are (unbalanced) binary classification problems for income prediction and loan assignment, respectively.
The datasets have both categorical and numerical features. Some of these features are actionable (e.g., \textit{occupation}, \textit{education}), while others are immutable (e.g., \textit{age}, \textit{sex}, \textit{native\_country}).
These datasets come without a causal graph and users' preferences over the features.
Following previous work \citep{karimi2021algorithmic, detoni2022synthesizing, naumann2021consequence}, we manually defined and fixed the cost correlation structures for both. We randomly generated user-specific weights for each instance by sampling from a (dataset-specific) mixture of Gaussians with $M=6$ components \GDT{and uniform mixture weights}.
\GDT{In \cref{sec:appendix-competitors-and-experimental-settings}, we give a description of the feature space, action space and the directed acyclic graphs needed by the CCS for both datasets.}
We then split the data into training ($70\%$), validation ($10\%$) and test ($20\%$) sets.
For each dataset, we designed the black-box classifier $h$ as an MLP with two hidden layers. We trained it by cross-entropy minimization, selecting the hyperparameters which maximise the $F_1$ score on the validation set.
In \cref{sec:pear-shallow-classifiers} we show additional experiments with shallow classifiers, such as a logistic model and a decision tree.

\textbf{``Easy'' vs.\@ ``Hard'' Users.}  Intuitively, users close to the decision boundary of the black-box $h$  will require few actions to achieve recourse, while users to whom $h$ assigns a low score might need longer and more complex interventions. Understanding the preferences of these ``hard'' users is crucial since a wrong suggestion might substantially increase the overall cost for them. For each dataset, we thus built two separate testing sets. The first one, named \texttt{All}, is obtained by sampling 300 users $\vs$ with an unfavourable classification ($h(\vs) < 0.5$), regardless of the actual value of $h$. The second one, named \texttt{Hard}, is obtained by sampling 300 users with an unfavourable classification having a score in the lower quartile of the black-box score distribution.

\textbf{Competitors. } We compare \method against several baselines: \fare and its explainable version \texttt{EFARE} \citep{detoni2022synthesizing}, \cscf \citep{naumann2021consequence}, an evolutionary algorithm which, similarly to \fare, generates recourse options by considering consequence-aware cost functions and action sets $\mathcal{A}$, and \face \citep{poyiadzi2020face}, a well-know AR algorithm, which optimizes for population-based  "feasible paths" to achieve recourse. We also consider two simpler baselines, a brute-force search (\texttt{MCTS}) and a vanilla reinforcement learning agent (\texttt{RL}), trained in a similar way as in \citep{hamrick2019combining, verma2022amortized}. Note that all the competitors are model-agnostic and \textit{not} interactive, since they assume the users' costs to be fixed. We provide more in-depth details on the baselines and their implementations in \cref{sec:appendix-competitors-and-experimental-settings}.

\textbf{Experimental Protocol.}  For \method, we vary the number of questions $T$ to the user from 0 to 10.
For $T=0$, we initialize the weights with the expected value of the prior, $\bbE_{P(\vw)}[\vw]$, that represents a user-independent population-based prior.  Moreover, we employ two user response models, the \textit{noiseless} model (\cref{eqn:noiseless}), to check the effectiveness of our approach in the best-case scenario where the user can perfectly express their preferences, and the \textit{logistic} model (\cref{eq:logistic}), to challenge our approach in a more realistic scenario.
To provide a fair comparison, we equip the competitors with our cost function (\cref{eqn:cost}) and set their weights to the expected value of the prior.

\begin{table}[]
\centering
\caption{Performance of all competitors averaged over $10$ runs.  A `-' indicates that the method did not find \textit{any} successful intervention for \textit{any} user. $\mathtt{PEAR}_{NL}$ and $\mathtt{PEAR}_{L}$ indicate \method associated with the noiseless and logistic response model, respectively. The best results are boldfaced.}
\label{tab:eval-results}
\resizebox{\textwidth}{!}{%
\begin{tabular}{@{}cccccccc@{}}
\toprule
 &  & \multicolumn{3}{c}{\texttt{Adult}} & \multicolumn{3}{c}{\texttt{GiveMeSomeCredit}} \\
\multirow{-2}{*}{\textbf{Users}} & \multirow{-2}{*}{\textbf{Method}} & \textit{Validity} & \textit{Cost} & \textit{Length} & \textit{Validity} & \textit{Cost} & \textit{Length} \\ \midrule
 & $\mathtt{FARE}$ & \cellcolor[HTML]{EFEFEF}\small{$0.90 \pm 0.27$} & \cellcolor[HTML]{EFEFEF}\small{$285.63 \pm 195.68$} & \cellcolor[HTML]{EFEFEF}\small{$3.45 \pm 1.19$} & \cellcolor[HTML]{EFEFEF}\small{$0.86 \pm 0.23$} & \cellcolor[HTML]{EFEFEF}\small{$161.77 \pm 107.42$} & \cellcolor[HTML]{EFEFEF}\small{$3.27 \pm 1.13$} \\
 & $\mathtt{CSCF}$ & \small{$0.78 \pm 0.28$} & \small{$154.28 \pm 125.34$} & \small{$2.53 \pm 0.57$} & \small{$0.57 \pm 0.42$} & \small{$100.69 \pm 120.22$} & \small{$2.51 \pm 1.12$} \\
 & $\mathtt{EFARE}$ & \cellcolor[HTML]{EFEFEF}\small{$0.76 \pm 0.39$} & \cellcolor[HTML]{EFEFEF}\small{$306.11 \pm 199.02$} & \cellcolor[HTML]{EFEFEF}\small{$3.54 \pm 1.25$} & \cellcolor[HTML]{EFEFEF}\small{$0.67 \pm 0.38$} & \cellcolor[HTML]{EFEFEF}\small{$155.92 \pm 109.19$} & \cellcolor[HTML]{EFEFEF}\small{$3.18 \pm 1.18$} \\
 & $\mathtt{RL}$ & \small{$0.76 \pm 0.38$} & \small{$283.31 \pm 167.86$} & \small{$3.36 \pm 1.08$} & \small{$0.12 \pm 0.32$} & \small{$59.66 \pm 0.00$} & \small{$2.00 \pm 0.00$} \\
 & $\mathtt{MCTS}$ & \cellcolor[HTML]{EFEFEF}\small{$0.44 \pm 0.44$} & \cellcolor[HTML]{EFEFEF}\small{$445.65 \pm 201.28$} & \cellcolor[HTML]{EFEFEF}\small{$4.60 \pm 1.19$} & \cellcolor[HTML]{EFEFEF}\small{$0.70 \pm 0.42$} & \cellcolor[HTML]{EFEFEF}\small{$214.33 \pm 119.38$} & \cellcolor[HTML]{EFEFEF}\small{$4.09 \pm 1.31$} \\
 & $\mathtt{FACE}$ & \small{$0.15 \pm 0.27$} & \small{$397.49 \pm 128.43$} & \small{$3.76 \pm 0.64$} & \small{$0.24 \pm 0.38$} & \small{$327.18 \pm 78.85$} & \small{$5.97 \pm 0.62$} \\[0.1cm]
 & $\mathtt{PEAR}_{NL}$ \textbf{(ours)} & \cellcolor[HTML]{EFEFEF}\small{$\bm{1.00 \pm 0.03}$} & \cellcolor[HTML]{EFEFEF}\small{$\bm{142.23 \pm 61.75}$} & \cellcolor[HTML]{EFEFEF}\small{$\bm{2.84 \pm 0.59}$} & \cellcolor[HTML]{EFEFEF}\small{$\bm{0.89 \pm 0.00}$} & \cellcolor[HTML]{EFEFEF}\small{$\bm{96.04 \pm 31.96}$} & \cellcolor[HTML]{EFEFEF}\small{$\bm{2.79 \pm 0.42}$} \\
\multirow{-8}{*}{\texttt{All}} & $\mathtt{PEAR}_{L}$ \textbf{(ours)} & \small{$\bm{1.00 \pm 0.04}$} & \small{$\bm{146.54 \pm 63.09}$} & \small{$\bm{2.84 \pm 0.56}$} & \small{$\bm{0.89 \pm 0.00}$} & \small{$\bm{100.19 \pm 29.53}$} & \small{$\bm{2.85 \pm 0.48}$} \\ \midrule
 & $\mathtt{FARE}$ & \small{$0.71 \pm 0.44$} & \small{$438.97 \pm 188.50$} & \small{$4.54 \pm 1.21$} & \small{$0.47 \pm 0.37$} & \small{$319.46 \pm 96.36$} & \small{$4.97 \pm 0.70$} \\
 & $\mathtt{EFARE}$ & \cellcolor[HTML]{EFEFEF}\small{$0.55 \pm 0.48$} & \cellcolor[HTML]{EFEFEF}\small{$454.05 \pm 202.76$} & \cellcolor[HTML]{EFEFEF}\small{$4.52 \pm 1.25$} & \cellcolor[HTML]{EFEFEF}\small{$0.22 \pm 0.36$} & \cellcolor[HTML]{EFEFEF}\small{$371.58 \pm 82.18$} & \cellcolor[HTML]{EFEFEF}\small{$5.31 \pm 0.71$} \\
 & $\mathtt{RL}$ & \small{$0.55 \pm 0.47$} & \small{$433.64 \pm 152.67$} & \small{$4.33 \pm 1.24$} & - & - & - \\
 & $\mathtt{CSCF}$ & \cellcolor[HTML]{EFEFEF}\small{$0.25 \pm 0.36$} & \cellcolor[HTML]{EFEFEF}\small{$382.84 \pm 126.31$} & \cellcolor[HTML]{EFEFEF}\small{$3.70 \pm 0.56$} & \cellcolor[HTML]{EFEFEF}\small{$0.13 \pm 0.32$} & \cellcolor[HTML]{EFEFEF}\small{$190.81 \pm 119.99$} & \cellcolor[HTML]{EFEFEF}\small{$3.36 \pm 1.11$} \\
 & $\mathtt{MCTS}$ & \small{$0.21 \pm 0.40$} & \small{$599.20 \pm 153.44$} & \small{$5.53 \pm 0.72$} & \small{$0.40 \pm 0.43$} & \small{$353.75 \pm 99.26$} & \small{$5.43 \pm 0.88$} \\
 & $\mathtt{FACE}$ & \cellcolor[HTML]{EFEFEF}\small{$0.00 \pm 0.04$} & \cellcolor[HTML]{EFEFEF}\small{$448.72 \pm 0.00$} & \cellcolor[HTML]{EFEFEF}\small{$5.20 \pm 0.00$} & \cellcolor[HTML]{EFEFEF}\small{$0.20 \pm 0.36$} & \cellcolor[HTML]{EFEFEF}\small{$455.20 \pm 92.10$} & \cellcolor[HTML]{EFEFEF}\small{$7.09 \pm 0.53$} \\[0.1cm]
 & $\mathtt{PEAR}_{NL}$ \textbf{(ours)} & \small{$\bm{0.99 \pm 0.08}$} & \small{$\bm{296.37 \pm 43.84}$} & \small{$\bm{3.35 \pm 0.55}$} & \small{$\bm{0.58 \pm 0.04}$} & \small{$\bm{251.60 \pm 51.16}$} & \small{$\bm{4.59 \pm 0.40}$} \\
\multirow{-8}{*}{\texttt{Hard}} & $\mathtt{PEAR}_{L}$ \textbf{(ours)} & \cellcolor[HTML]{EFEFEF}\small{$\bm{0.99 \pm 0.09}$} & \cellcolor[HTML]{EFEFEF}\small{$\bm{301.13 \pm 52.61}$} & \cellcolor[HTML]{EFEFEF}\small{$\bm{3.34 \pm 0.58}$} & \cellcolor[HTML]{EFEFEF}\small{$\bm{0.58 \pm 0.02}$} & \cellcolor[HTML]{EFEFEF}\small{$\bm{262.23 \pm 45.36}$} & \cellcolor[HTML]{EFEFEF}\small{$\bm{4.64 \pm 0.36}$} \\ \bottomrule
\end{tabular}%
}
\end{table}

\textbf{Q1: \method successfully minimizes the regret.} \cref{fig:average-regret} shows the evaluation of the regret as a function of the number of queries to the user. Here the ground-truth intervention $I^{GT}$ (which is unknown) is approximated by running \method with the correct user costs $\vw^{GT}$, and the regret is normalized by rescaling the costs between $C(I^\textrm{GT}\mid \vw^{GT})$ and $C(I^{(0)} \mid \vw^{GT})$ where we generate $I^{(0)}$ using the expectation of the prior.
We run \method with two different choice set dimensions, $k=2$ and $k=4$, and for both noiseless and logistic response models. After a few questions, \method reaches a low regret in all settings. Generally, a larger choice set produces a lower regret, irrespective of the response model, with the downside of increasing the cognitive burden for the user. We now briefly summarize the results when $T=10$. For the \texttt{Adult} dataset, the best regret is $\approx 0.09$ for the noiseless user and $k=4$, while the worst regret is $\approx 0.40$ for the logistic response model and $k=2$. For \texttt{GiveMeSomeCredit}, we get $\approx 0.15$ (noiseless, $k=4$) and $\approx 0.45$ (logistic, $k=2$). Overall, we can provide interventions which are at least $50\%$ cheaper than their preference-agnostic counterparts.

\begin{table}[t]
\centering
\caption{Evaluation of \method (with $q=10$ and a logistic noise model) for an increasing amount of CCS graph corruption, averaged over 10 runs. "None" indicates that the correct causal graph is being used.}
\resizebox{\textwidth}{!}{%
\begin{tabular}{@{}cccccccc@{}}
\toprule
 &  & \multicolumn{3}{c}{Adult} & \multicolumn{3}{c}{GiveMeSomeCredit} \\
\multirow{-2}{*}{\textbf{Users}} & \multirow{-2}{*}{\textbf{Corruption}} & \textit{Validity} & \textit{Cost} & \textit{Length} & \textit{Validity} & \textit{Cost} & \textit{Length} \\ \midrule
 & \textbf{None} & \cellcolor[HTML]{EFEFEF}\small{$1.00 \pm 0.04$} & \cellcolor[HTML]{EFEFEF}\small{$146.54 \pm 63.09$} & \cellcolor[HTML]{EFEFEF}\small{$2.84 \pm 0.56$} & \cellcolor[HTML]{EFEFEF}\small{$0.89 \pm 0.00$} & \cellcolor[HTML]{EFEFEF}\small{$100.19 \pm 29.53$} & \cellcolor[HTML]{EFEFEF}\small{$2.85 \pm 0.48$} \\
 & \textbf{0.15} & \small{$1.00 \pm 0.00$} & \small{$165.11 \pm 46.19$} & \small{$2.79 \pm 0.24$} & \small{$0.90 \pm 0.00$} & \small{$98.02 \pm 18.37$} & \small{$2.46 \pm 0.16$} \\
 & \textbf{0.25} & \cellcolor[HTML]{EFEFEF}\small{$1.00 \pm 0.00$} & \cellcolor[HTML]{EFEFEF}\small{$162.05 \pm 48.05$} & \cellcolor[HTML]{EFEFEF}\small{$2.89 \pm 0.34$} & \cellcolor[HTML]{EFEFEF}\small{$0.90 \pm 0.07$} & \cellcolor[HTML]{EFEFEF}\small{$99.51 \pm 18.53$} & \cellcolor[HTML]{EFEFEF}\small{$2.48 \pm 0.18$} \\
 & \textbf{0.5} & \small{$1.00 \pm 0.00$} & \small{$175.54 \pm 62.29$} & \small{$2.82 \pm 0.36$} & \small{$0.90 \pm 0.11$} & \small{$110.35 \pm 19.93$} & \small{$2.56 \pm 0.20$} \\
\multirow{-5}{*}{All} & \textbf{1.0} & \cellcolor[HTML]{EFEFEF}\small{$0.99 \pm 0.02$} & \cellcolor[HTML]{EFEFEF}\small{$172.51 \pm 61.13$} & \cellcolor[HTML]{EFEFEF}\small{$2.96 \pm 0.40$} & \cellcolor[HTML]{EFEFEF}\small{$0.91 \pm 0.04$} & \cellcolor[HTML]{EFEFEF}\small{$106.63 \pm 28.84$} & \cellcolor[HTML]{EFEFEF}\small{$2.64 \pm 0.28$} \\ \midrule
 & \textbf{None} & \small{$0.99 \pm 0.09$} & \small{$301.13 \pm 52.61$} & \small{$3.34 \pm 0.58$} & \small{$0.58 \pm 0.02$} & \small{$262.23 \pm 45.36$} & \small{$4.64 \pm 0.36$} \\
 & \textbf{0.15} & \cellcolor[HTML]{EFEFEF}\small{$1.00 \pm 0.00$} & \cellcolor[HTML]{EFEFEF}\small{$308.22 \pm 38.40$} & \cellcolor[HTML]{EFEFEF}\small{$3.47 \pm 0.26$} & \cellcolor[HTML]{EFEFEF}\small{$0.63 \pm 0.04$} & \cellcolor[HTML]{EFEFEF}\small{$237.00 \pm 35.72$} & \cellcolor[HTML]{EFEFEF}\small{$4.06 \pm 0.31$} \\
 & \textbf{0.25} & \small{$1.00 \pm 0.02$} & \small{$305.62 \pm 39.73$} & \small{$3.32 \pm 0.35$} & \small{$0.66 \pm 0.11$} & \small{$238.92 \pm 28.66$} & \small{$3.90 \pm 0.27$} \\
 & \textbf{0.5} & \cellcolor[HTML]{EFEFEF}\small{$1.00 \pm 0.03$} & \cellcolor[HTML]{EFEFEF}\small{$314.36 \pm 37.93$} & \cellcolor[HTML]{EFEFEF}\small{$3.38 \pm 0.32$} & \cellcolor[HTML]{EFEFEF}\small{$0.59 \pm 0.14$} & \cellcolor[HTML]{EFEFEF}\small{$250.47 \pm 29.71$} & \cellcolor[HTML]{EFEFEF}\small{$4.16 \pm 0.24$} \\
\multirow{-5}{*}{Hard} & \textbf{1.0} & \small{$0.99 \pm 0.03$} & \small{$313.51 \pm 61.64$} & \small{$3.69 \pm 0.44$} & \small{$0.64 \pm 0.09$} & \small{$256.24 \pm 12.04$} & \small{$4.24 \pm 0.18$} \\ \bottomrule
\end{tabular}%
}
\label{tab:error-causal}
\end{table}

\textbf{Q2: \method outperforms competitors in terms of validity and cost.}  Following the AR literature \citep{karimi2020survey, verma2020counterfactual}, we compare \method (with $T=10$) and all competitors in terms of average \textit{validity}, \ie fraction of users for which we obtained recourse, intervention \textit{cost} and \textit{length} (or \textit{sparsity}), \ie the number of features that have to be changed.
Intervention costs are computed by using the true weights $\vw^{GT}$. \cref{tab:eval-results} shows the results.  \method manages to achieve the highest validity while also providing substantially cheaper interventions than the non-personalized competitors on average. This is true both for the noiseless and logistic response models.
While \cscf tends to produce shorter interventions, these are in general more costly and have a larger cost variance with respect to those found by \method, confirming the intuition that length is a suboptimal proxy of intervention complexity. The only exception is the \texttt{Hard} setting of \texttt{GiveMeSomeCredit}, where however \cscf manages to achieve recourse for only 22\% of the users, whereas \method achieves recourse in 58\% of the cases. The difficulty of \cscf in achieving recourse is visible in all settings and severely limits its applicability. Furthermore, \cscf is 10 to 50 times more computationally expensive than \method, making it unsuitable for real-time interactive scenarios.
The \mcts baseline has rather poor performance both in terms of validity and cost in all settings, while
the \rl baseline has a reasonably high validity on \texttt{Adult} but it completely fails to learn a policy achieving recourse on \texttt{GiveMeSomeCredit}. On the other hand, methods which combine \mcts and \rl (\fare and \texttt{EFARE}) give better performance, which is aligned with previous results \citep{detoni2022synthesizing}, but are still suboptimal with respect to \method in terms of both validity and cost.
Finally, \face struggles to achieve recourse since it needs to find a "feasible path" from the current user to a similar one \textit{in the training set}, which is favourably classified.

\textbf{Q3: \method is robust to misspecifications of the cost correlation structure.} In the previous experiments, following other research works \citep{karimi2021algorithmic, detoni2022synthesizing, naumann2021consequence}, we assumed to know the structure of the CCS a-priori. However, in a real scenario, we might have instead an \textit{approximate} causal graph from which to derive the CCS.
\cref{tab:error-causal} shows the validity, cost and length of the interventions found by removing $X\%$ of edges from the causal graph, with $X \in \{0.15, 0.25, 0.50, 1.00\}$.
Validity is almost unaffected by corruption in all settings since it only impacts the computation of the cost. On the other hand, as expected, increasing the amount of graph corruption reduces the effectiveness of user feedback. However, the degradation is not dramatic. Indeed, if we look at the \texttt{Hard} evaluation, the increase in cost is negligible (around 4\%) with up to $50\%$ randomly removed edges.
On \texttt{GiveMeSomeCredit}, we do not see any significant increase in costs. Surprisingly, we see instead an improvement for $15\%$ and $25\%$ corruption levels. We hypothesize that lacking causal knowledge about features which are not needed for recourse can be beneficial since it simplifies the elicitation process. However, at higher levels of corruption, this effect disappears.
The setting $X=1.0$ is equivalent to a non-causal cost function, in which acting on a feature has always the same cost, irrespective of the others. It is the common choice of many works dealing with AR \citep{wachter2017counterfactual, ramakrishnan2020synthesizing, poyiadzi2020face}.
Under such a setting, when considering \texttt{All} users, the degradation is more evident, but still within $6\%$ for \texttt{GiveMeSomeCredit}, while for \texttt{Adult} it goes up to $15\%$.
Overall, results clearly indicate that \method can suggest reasonable cost interventions even with a largely misspecified cost correlation structure. This is apparent when comparing these results with those in \cref{tab:eval-results}. Even with $50\%$ randomly removed edges, \method recommends interventions that are cheaper than all competitors but \cscf, that however has a substantially lower validity.

\textbf{Further experimental evaluations.} We performed some additional experiments which explored the relationship between the cost and length of an intervention $I$ and validated the usage of the model score $h(\vx)$ as a proxy for identifying the difficulty of finding a recourse suggestion. We report our findings in \cref{sec:appendix-further-evaluations}. Lastly, following the example of \citet{detoni2022synthesizing}, we also provide an explainable version of our method, \xmethod, which complements interventions with Boolean explanations during the elicitation process by taking into consideration the user effort. We then compared \xmethod against the corresponding baselines \texttt{EFARE} \citep{detoni2022synthesizing} and while it provides good performances at the expense of being more interpretable, it is still outperformed by \method. The results are presented in \cref{sec:user-aware-explainable-interventions}.

\section{\changed{Broader Impact}}

\changed{
Algorithmic recourse focuses on developing methods to increase the user's agency in contesting automated decision systems outcomes and the fairness of the current machine learning models. As for all systems impacting humans, we need to consider the possible bad ethical ramifications of these technologies and potential mitigation strategies. In the context of personalized algorithmic recourse systems, we can identify two sources of algorithmic bias which could hinder the quality of our recommendations:}

\changed{\textbf{(1) Different recourse suggestions when conditioned on a protected attribute}. Users who share a similar profile, but differ in some sensitive features (e.g., sex, age, etc.) might obtain different interventions $I$. We believe this might be the most important source of unfairness. In such a scenario, a method might successfully optimize both \cref{eqn:par} and \cref{eqn:par_regret}, but some categories of users will always get costlier interventions. However, this is an issue shared by almost the totality of methods in the AR literature.}

\changed{\textbf{(2) Uninformative prior lacking information about sensitive categories}. The prior $P(\vw)$ used by the Bayesian procedure in \method might not describe the preferences of less-represented groups, thus making it more difficult to learn such cost functions. Critically, methods employing a population-level estimate of $\vw$ or which relies on expert evaluation \citep{rawal2020beyond} have the same limitation.}

\changed{We believe issue (1) relates to the quality underlying recourse generator method (e.g., \texttt{W-FARE}, \texttt{CSCF}, \texttt{FACE}, etc.). A solution could be employing AR strategies that are specifically optimized to satisfy a notion of fairness \citep{gupta2019equalizing, haldar2022raguel, von2022fairness}. However, to the best of our knowledge, this is a little-explored area in the recourse literature which would require additional investigation beyond the scope of the present work. Regarding issue (2), we could imagine focusing our efforts on collecting additional data on sensitive groups to learn a more informative prior to running \method.
Additionally, we could raise the iteration budget ($T$) to increase the number of pairwise constraints we apply to our MCMC procedure. However, it would also increase the cognitive burden for the user. }

Lastly, eliciting users’ preferences might entail asking sensitive questions, or malicious entities could exploit these procedures to "hack" and twist the intervention generation via strategic behaviour \citep{hardt2016strategic}. These considerations can be mitigated by research on adversarial attacks to ensure the method’s robustness \citep{chen2020strategic}. Moreover, legal advice might be needed to manage personal user data.

\section{Conclusion}

In this work, we identify the problem of \textit{personalized} algorithmic recourse as a fundamental stepping stone for ensuring recourse is usable in real-world applications, and develop \method, the first algorithm able to provide \textit{personalized} interventions. Our experimental evaluation shows that \method substantially outperforms existing (non-personalized) solutions in terms of both validity and intervention cost with only a handful of queries to the user.
We hope that this initial contribution can foster further research in the community to work towards a more realistic form of algorithmic recourse that can be successfully deployed in real-world scenarios.
As for all methods dealing with algorithmic recourse, the effectiveness of the approach should, in principle, be evaluated on real users. However, this evaluation is highly non-trivial (and thus still missing in the algorithmic recourse literature) because it requires the creation of a realistic scenario where a user feels to be \textit{unfairly treated} in some machine-driven decision involving her life. The legal requirements that are progressively being introduced to regulate AI systems~\citep{voigt2017eu} could contribute to making the information needed to set up such a scenario available in the near future.

\section*{Acknowledgments}

Funded by the European Union. Views and opinions expressed are however those of the author(s) only and do not necessarily reflect those of the European Union or the European Health and Digital Executive Agency (HaDEA). Neither the European Union nor the granting authority can be held responsible for them. Grant Agreement no. 101120763 - TANGO. AP and BL also acknowledge the support of the MUR PNRR project FAIR - Future AI Research (PE00000013) funded by the NextGenerationEU. BL also acknowledges the support of the European Commission under Horizon Europe Programme, grant number 101120237 - ELIAS. The work of Giovanni De Toni was partially supported by the project AI@Trento (FBK-Unitn).

\bibliographystyle{tmlr}
\bibliography{references}

\newpage
\appendix

\section{Competitors and Experimental Settings}
\label{sec:appendix-competitors-and-experimental-settings}

We now briefly illustrate the competitors' main principles and describe their most important hyperparameters.  We also discuss \WFARE and its new reward function in more detail.

\textbf{Implementation details.} We implemented \method, the competitors and the black box classifiers using Python (>= 3.7) and PyTorch \citep{paszke2019pytorch}. For reproducibility purposes, the code and the pre-trained models are freely available online\footnote{\url{https://github.com/unitn-sml/pear-personalized-algorithmic-recourse}}. We used the original code for both \fare\footnote{\url{https://github.com/unitn-sml/syn-interventions-algorithmic-recourse}} and \cscf\footnote{\url{https://github.com/ppnaumann/CSCF}}, with minimal modifications to make them compatible with our experimental setting. For \face, we used the implementation available in the CARLA library \citep{pawelczyk2021carla}. As for the baselines, we adapted \mcts and the \rl agent from the \fare repository. All the experiments were run on a virtual machine running CentOS 7.6.18 with 165 cores, and 25 GiB of RAM. The full set of hyperparameter configurations is available in the code.

\GDT{
\textbf{State space and Action space.} Given the raw datasets, we one-hot encoded categorical features and we performed min-max normalization for the continuous features using \texttt{scikit-learn} \citep{scikit-learn}. We also manually performed additional standard data engineering tasks, such as removing entries with null values or checking for potential outliers. After the data cleaning and preprocessing steps, we kept the following features for each dataset:
\begin{itemize}
    \item \texttt{Adult}: \textit{age}, \textit{capital\_gain}, \textit{capital\_loss}, \textit{hours\_per\_week} (continuous) \textit{workclass}, \textit{education}, \textit{marital\_status}, \textit{occupation}, \textit{relationship}, \textit{race}, \textit{sex},  \textit{native\_country} (categorical). We considered the features \textit{age}, \textit{relationship}, \textit{race}, \textit{sex}, and \textit{native\_country} to be unactionable.
    \item \texttt{GiveMeSomeCredit}: \textit{RevolvingUtilizationOfUnsecuredLines}, \textit{age}, \textit{NumberOfTime30-59DaysPastDueNotWorse}, \textit{DebtRatio}, \textit{MonthlyIncome}, \textit{NumberOfOpenCreditLinesAndLoans}, \textit{NumberOfTimes90DaysLate}, \textit{NumberRealEstateLoansOrLines},\textit{NumberOfTime60-89DaysPastDueNotWorse}, \textit{NumberOfDependents} (continuous). We considered the features \textit{age} and \textit{NumberOfDependents} to be unactionable.
\end{itemize}
These features represent the state $\vs \in \calS$ of a user.  
\cref{tab:action-space-experiment} reports the action space $\calA$ used by \texttt{W-FARE} and the baselines for each of the experimental datasets. For \texttt{Adult}, we adopted the same action set used by \cite{detoni2022synthesizing}, while for \texttt{GiveMeSomeCredit} we devised the functions ourselves. Please note that the actions $a \in \calA$ operate directly in the original feature space, without encoding or rescaling, to ensure the interventions are understandable by humans. The features assigned as non-actionable do not have any action available. We refer the reader to the code implementation for additional details. 
} 

\begin{figure}[t]
    \centering
    \begin{subfigure}{0.5\linewidth}
        \includegraphics[width=\linewidth]{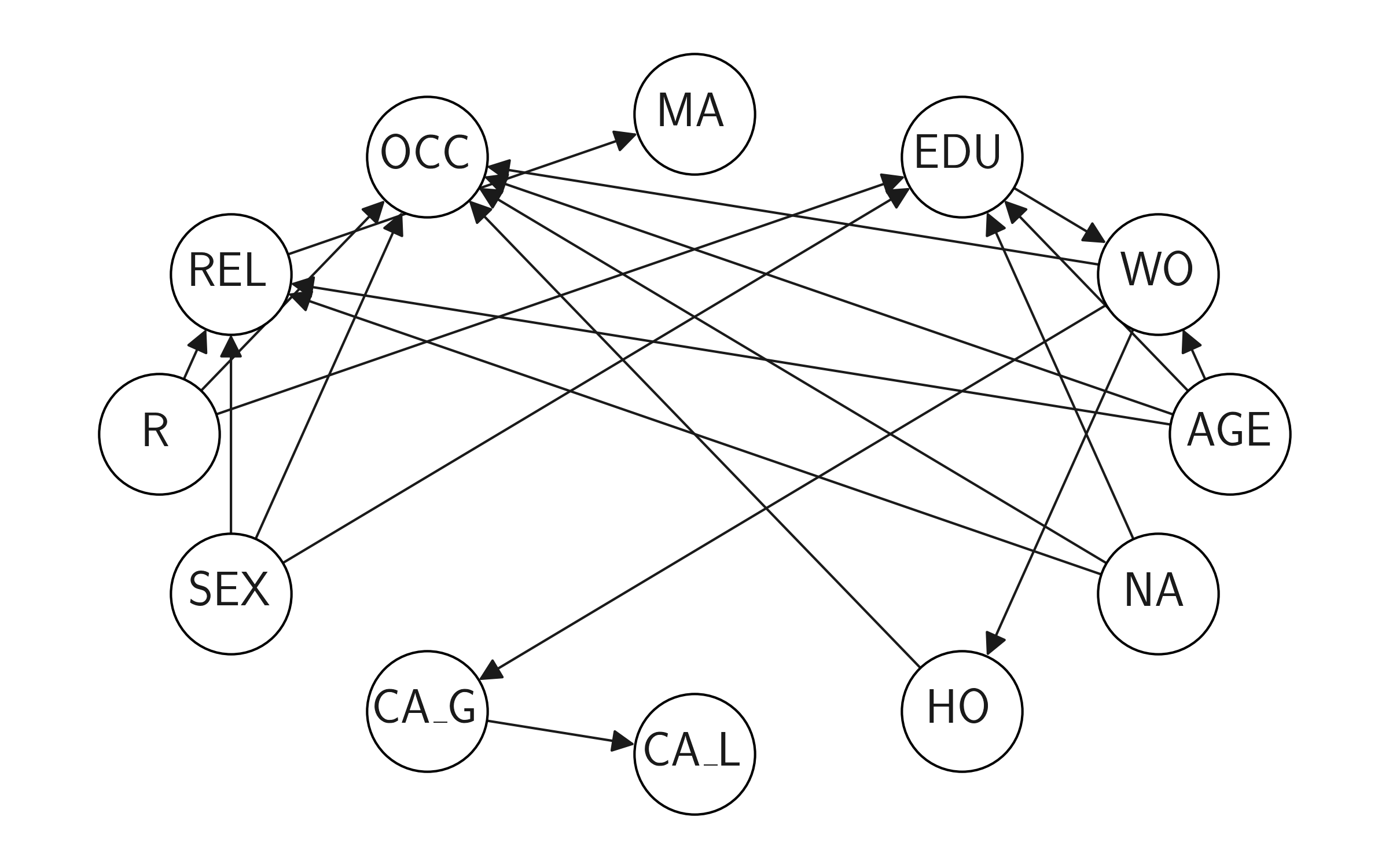}
        \caption{\texttt{Adult}}
    \end{subfigure}%
    \begin{subfigure}{0.5\linewidth}
        \includegraphics[width=\linewidth]{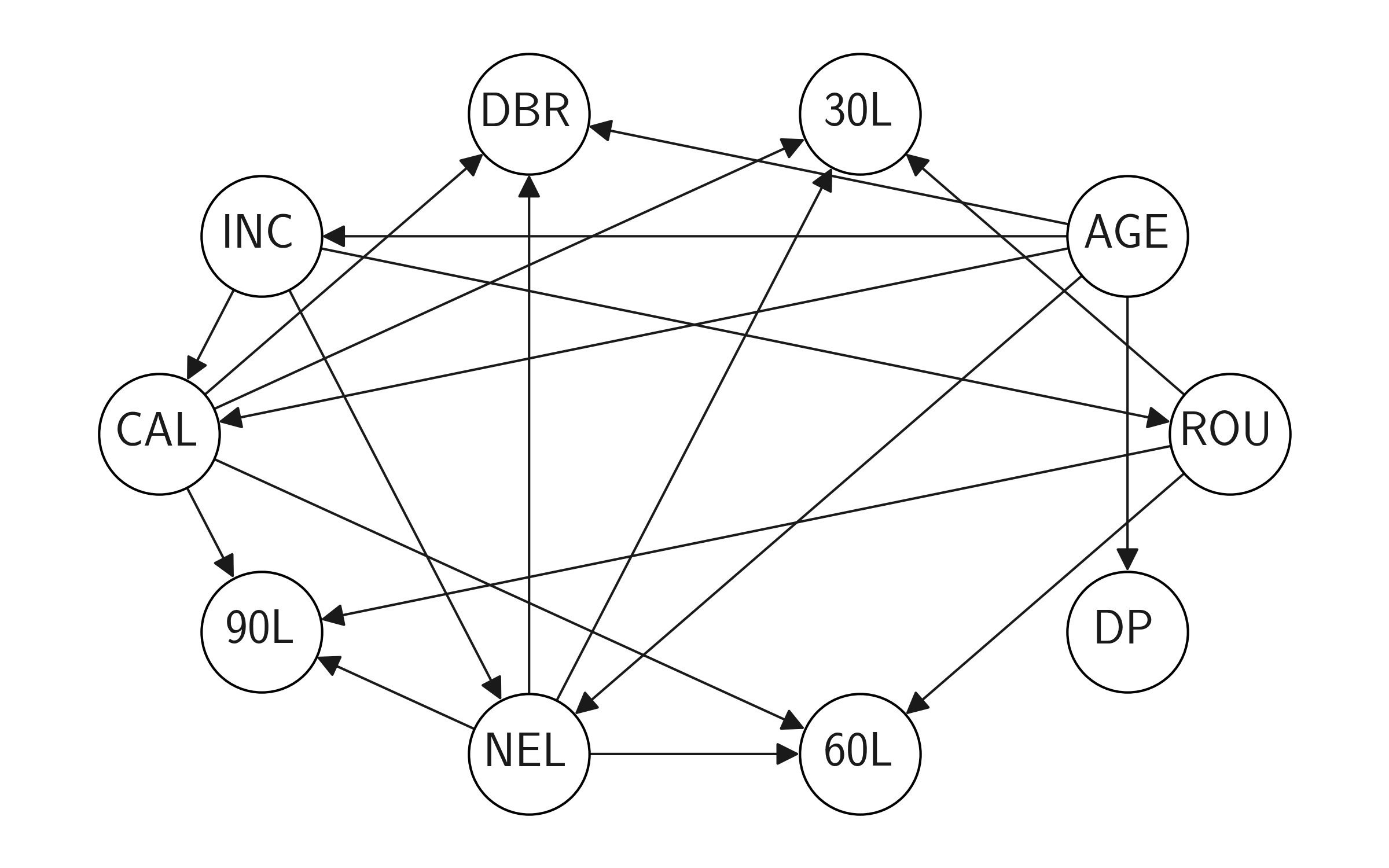}
        \caption{\texttt{GiveMeSomeCredit}}
    \end{subfigure}
    \caption{\textbf{Directed Acyclic Graphs used for the Cost Correlation Structures.} Each node corresponds to a specific user feature. The complete mapping between node acronyms and features can be found in the experimental configuration files.}
    \label{fig:causal-graphs}
\end{figure}

\textbf{Cost Correlation Structure. } Following the previous literature, \citep{karimi2021algorithmic, detoni2022synthesizing, naumann2021consequence}, we manually designed the directed acyclic graph for both experiments. \cref{fig:causal-graphs} shows the resulting graphs. During the graph corruption experiments, we incrementally remove a subset of edges taken uniformly at random. 

\begin{table}[p!]
\centering
\caption{\GDT{Action set $\calA$ for \texttt{Adult} and \texttt{GiveMeSomeCredit} datasets used by \texttt{W-FARE} and the baselines. We report the functions, their arguments and which feature each of them modifies. For \texttt{Adult}, we adopted the same action set as \cite{detoni2022synthesizing}, but we added two additional functions, CHANGE\_CAP\_LOSS and CHANGE\_CAP\_GAIN to increase the available interventions. We manually crafted the action space for \texttt{GiveMeSomeCredit}. In the case of continuous features, we constructed the argument sets by picking evenly spaced numbers from an interval since MCTS cannot handle continuous search spaces. We report the intervals from which we selected the argument values. For categorical features, we simply report the full set of options available.}}
\label{tab:action-space-experiment}
\subcaption*{\texttt{Adult}}
\resizebox{\textwidth}{!}{%
\begin{tabular}{@{}lllp{7cm}@{}}
\toprule
\textbf{Feature} & \multicolumn{1}{c}{\textbf{Function ($f$)}} & \multicolumn{1}{c}{\textbf{Argument Type}} & \textbf{Argument ($x$)}                                                                                                                                                                                                  \\ \midrule
workclass        & CHANGE\_WORKCLASS                           & Categorical                                & Never-worked, Without-pay, Self-emp-not-inc, Self-emp-inc, Private, Local-gov, State-gov, Federal-gov                                                                                                                    \\
education        & CHANGE\_EDUCATION                           & Categorical                                & Preschool, 1st-4th, 5th-6th, 7th-8th, 9th, 10th, 11th, 12th, HS-grad, Some-college, Bachelors, Masters, Doctorate, Assoc-acdm, Assoc-voc, Prof-school                                                                    \\
occupation       & CHANGE\_OCCUPATION                          & Categorical                                & Tech-support, Craft-repair, Other-service, Sales, Exec-managerial, Prof-specialty, Handlers-cleaners, Machine-op-inspct, Adm-clerical, Farming-fishing, Transport-moving, Priv-house-serv, Protective-serv, Armed-Forces \\
capital\_gain    & CHANGE\_CAP\_GAIN                           & Integer                                   & $[-5000, -1] \cup [1, 5000]$                                                                                                                                                                                             \\
capital\_loss    & CHANGE\_CAP\_LOSS                           & Integer                                  & $[-5000, -1] \cup [1, 5000]$                                                                                                                                                                                             \\
hours\_per\_week & CHANGE\_HOURS                               & Integer                                   & $[-25, -1] \cup [1, 25]$                                                                                                                                                                                                 \\ \bottomrule
\end{tabular}%
}
\bigskip
\subcaption*{\texttt{GiveMeSomeCredit}}
\resizebox{\textwidth}{!}{%
\begin{tabular}{@{}llll@{}}
\toprule
\multicolumn{1}{c}{\textbf{Feature}} & \multicolumn{1}{c}{\textbf{Function} ($f$)} & \multicolumn{1}{c}{\textbf{Argument Type}} & \multicolumn{1}{c}{\textbf{Argument} (x)} \\ \midrule
RevolvingUtilizationOfUnsecuredLines & BALANCE                               & Real                                       & $[-1, -0.01] \cup [0.01, 1]$          \\
NumberOfTime30-59DaysPastDueNotWorse & PAST\_DUE\_30                         & Integer                                    & $[-25, -1]$                           \\
DebtRatio                            & DEBT\_RATIO                           & Real                                       & $[-1, -0.01] \cup [0.01, 1]$          \\
MonthlyIncome                        & INCOME                                & Integer                                    & $[-10000, -1] \cup [1, 10000]$        \\
NumberOfOpenCreditLinesAndLoans      & OPEN\_LOANS                           & Integer                                    & $[-25, -1]$                           \\
NumberOfTimes90DaysLate              & PAST\_DUE\_90                         & Integer                                    & $[-25, -1]$                           \\
NumberRealEstateLoansOrLines         & OPEN\_REAL\_ESTATE                    & Integer                                    & $[-25, -1]$                           \\
NumberOfTime60-89DaysPastDueNotWorse & PAST\_DUE\_60                         & Integer                                    & $[-25, -1]$                           \\ \bottomrule
\end{tabular}%
}
\end{table}

\subsection{Monte Carlo Tree Search (\mcts)}

\mcts \citep{kocsis2006bandit, coulom2007efficient} is an efficient tree-based heuristic search procedure that can solve difficult tasks where computing directly the exact solution is intractable \citep{silver2016mastering}. 
\mcts builds a tree representing all possible moves and outcomes.
Each tree node represents a state $\vs^{(i)}$ and each edge represents performing an action $a \in \mathcal{A}$.
In each iteration, \mcts builds this tree following four main steps: selection, expansion, simulation, and backpropagation.
The \textit{selection} step takes a state $\vs^{(t)}$ and selects the next action $a^{(t+1)}$, based on some strategy such as UCB1 \citep{kocsis2006bandit}. In our experiments, \mcts uses the P-UCT \citep{rosin2011multi, silver2016mastering} criterion, defined as:
\begin{equation}
    a^{(t+1)} = \argmax_{a' \in \mathcal{A}} \; \bbE[r | \vs^{(t)}, a'] + U(\vs^{(t)}, a') + L(\vs^{(t)}, a')
\end{equation}
Here, $\bbE[r | \vs^{t}, a']$ is the expected reward of taking action $a'$ in state $\vs^{t}$. The reward $r$ is $\gamma^{(|I|)}$ if we achieve recourse and $-1$ otherwise. $|I|$ is the length of the intervention while $\gamma \in (0, 1]$ is a discount factor. $L(\vs^{(t)}, a')$ is a penalty is equal to $c_{act\_cost} \exp(-C(a', \vs^{(t)} \mid \vw))$. $ U(\vs^{(t)}, a')$ trades off exploration and exploitation, and it is defined as
\begin{equation}
    U(\vs^{(t)}, a) = c_{puct} \cdot P(a  \mid \vs^{(t)}) \frac{\sqrt{N_{\vs^{(t)}}}}{1+N_{\vs^{(t+1)}}}
\end{equation}
where $P(a  \mid \vs^{(t)})$ is the prior probability of choosing such action, $N_{\vs^{(t)}}$ is the visit count of node $\vs^{(t)}$, and $N_{\vs^{(t+1)}}$ is the visit count of the child node being evaluated. In our experiments, the prior $ P(a  \mid \vs^{(t)})$ is a uniform distribution over the actions. 

The \textit{expansion} step adds new child nodes to the current node based on the available actions.
The \textit{simulation} step then carries out a \textit{playout}, selecting moves randomly until a terminal state is reached. The value of the outcome of this terminal state is then backpropagated up the tree (\textit{backpropagation} step), incrementing the visit $N_{\vs^{(i)}}$ and reward counts of nodes along the path that was traversed to reach the outcome. We use the reward counts to compute $\bbE[r | \vs^{(t)}, a']$. The \textit{number of simulations} is a hyperparameter. Once the simulations are done, we start over from the \textit{selection} step.
\mcts keeps expanding and selecting nodes until we reach recourse, or a user-defined \textit{maximum intervention length}. Once \mcts ends, we return the successful intervention to the user. An intervention here is a \textit{path} from the root node ($\vs^{(0)}$) to a terminal node ($\vs^{(T)}$).

\textbf{Hyperparameters.} In our experiments, we set the \textit{number of simulations} to 15 and 10 for \texttt{Adult} and \texttt{GiveMeSomeCredit}, respectively. We also set the \textit{maximum intervention length} to 6 and 8, for \texttt{Adult} and \texttt{GiveMeSomeCredit}, respectively. The value $c_{act\_cost}$ and $c_{puct}$ are also hyperparameters. We set them to $1$ and $0.5$ respectively, for both experiments.

\subsection{Recourse Policy (\rl)}

\rl is a simple agent which learns a recourse policy. Following the architecture described in \cite{detoni2022synthesizing}, it has four different components: an encoder $g_{enc}$, an LSTM controller $f_c$, a function network $g_f$, an argument network $g_x$ and a value network $g_V$. $g_{enc}$ produces a latent representation of the current state $\vs^{(t)}$ which is sent to the controller $f_c$. Given the hidden states $h_t$ coming from $f_c$, the function and argument network generates the policies $\pi_f$ and $\pi_x$ which we use to get the tuple $(f,x)^{(t+1)}$ which corresponds to the next action $a$. We use the value network $g_V$ to approximate the value function $V(\vs^{(t)}, a)$.

\textbf{Training the agent.} We train \rl such to minimize the following loss function
\begin{equation}
    \ell = \sum_{batch} (V-r^2)-{\pi_{f}^{true}}^T\log(\pi_f) - {\pi^{true}_x}^T\log(\pi_x)
    \label{eqn:rl-policy-loss}
\end{equation}
where we assume to have access to the ground truth policies ${\pi_{f}^{true}}$ and ${\pi_{x}^{true}}$ either by having a previous dataset of recourse options, by optimizing over an MDP following \citep{verma2022amortized}, or by discovering such traces by direct search. In our paper, we follow the last approach by using \mcts as the search component.
At inference time, we run the agent until we either find recourse, reach the \textit{maximum intervention length}, or call an illegal action (e.g., an action whose \textit{precondition} is not satisfied). 

\textbf{Hyperparameters.} As in \cite{detoni2022synthesizing}, $g_{enc}$, $g_f$ and $g_x$ are all MLPs and we manually defined their structure for each experimental setting (e.g., number of layers, dimensions, etc.). We do the same for the controller $f_c$. We optimize \cref{eqn:rl-policy-loss} via Adam and we set the \textit{learning rate} to 0.001 for \texttt{Adult}, and 0.003 for \texttt{GiveMeSomeCredit}.

\subsection{(Explainable) eFficient counterfActual REcourse (\fare and \EFARE)}
\label{sec:fare-efare}

\fare \citep{detoni2022synthesizing} is a model-agnostic method which exploits the synergy between a learned (\rl) policy and a discrete search procedure (\mcts) to discover recourse options efficiently.
At training time, differently from \mcts and \rl, we do not use a uniform prior, but we leverage the agent instead $P(a  \mid \vs^{(t)}) = \pi_f(\vs^{(t)})\cdot\pi_x(\vs^{(t)})$. Similarly to \citet{silver2016mastering}, we foster additional exploration by replacing $P(a  \mid \vs^{(t)})$ with a linear combination, $\epsilon_P P(a \mid \vs^{(t)}) + (1-\epsilon_P) \eta_P$ where $\eta_P \sim \hbox{Dir}(\alpha_d)$. We manually set the parameters $\alpha_d$ and $\epsilon_P$ for each setting.

\EFARE is a deterministic model which provides sequential counterfactuals together with explanations. Given a trained \fare model, a sample of successful interventions for the training set users is extracted. 
These \textit{interventions} are merged to form a graph, where each node represents an action $a$, and each arch represents which actions $(f,x)$ can be taken from the said node. 
Then, for each node, a small decision tree is trained to indicate, given a user state $\vs^{(t)}$, which is the next recommended action. 
Given this trained automaton, we can traverse the graph using the decision trees to transition between nodes. Moreover, the decision trees can give us Boolean explanations for each transition.

\textbf{Hyperparameters.} \fare shares the same hyperparameters of \mcts and \rl. During training, we set the \textit{number of simulations} to 15 and 10, for \texttt{Adult} and \texttt{GiveMeSomeCredit}, respectively. The noise fraction is instead set to $\epsilon_P=0.3$ for both, with $\eta_p = 0.3$. At inference time, we add no noise ($\epsilon_P=0$) and the \textit{number of simulations} is fixed at 5. Lastly, for each dataset, we train a corresponding \EFARE version by sampling 300 interventions from the training set using \fare.
 
\subsection{Consequence-aware Sequential Counterfactuals (\cscf)}

\cscf \citep{naumann2021consequence} is a model-agnostic evolutionary algorithm which can generate sequential recourse plans by taking into consideration feature relationships. \cscf uses the Biased Random-Key Genetic Algorithm (BRKGA) \citep{gonccalves2011biased} together with non-dominated sorting (NDS) \citep{srinivas1994muiltiobjective} to perform multi-objective optimization. BRKGA optimizes over the solution \textit{genotype} (real-values vectors $G \in [0,1]$) and evaluates its \textit{phenotype} instead (sequences of actions $a \in \mathcal{A}$). Thus, they propose also a \textit{decoder} to deterministically derive a single phenotype from a genotype.

At each iteration, \cscf evaluates the decoded solutions phenotypes by computing their \textit{fitness}. We use the same fitness function as proposed in the original paper. Then, following NDS, a new population is formed by genetic mating and biased crossover between the \textit{elites} (feasible and valid solutions in the Pareto front) and \textit{non-elites}. The loop then continues until we reach a termination condition (e.g., we reach the maximum number of generations). From the last population set, we pick the lowest cost valid intervention with ties broken uniformly at random.  

\textbf{Hyperparameters. } In our experiments, we mainly used the default hyperparameters provided by the authors \citep{naumann2021consequence}. We set the population size, $p=50$, and the maximum number of generations, $n=25$, for both \texttt{Adult} and \texttt{GiveMeSomeCredit}, to keep the computation time manageable. Higher values might improve the performances, but it would make \cscf unusable in an online setting (e.g., web interface provided by the bank).  

\subsection{Feasible and Actionable Counterfactual Explanations (\face)}

\face \citep{poyiadzi2020face} is a model-agnostic method which can find actionable counterfactual examples respecting the underlying data distribution. The main idea is to propose recourse options based on the shortest path between the current state of the user and a state reaching recourse. \face proposes different density-weighted metrics to find these "feasible paths". \face starts by building a graph over the training set data points. Then, the user can specify certain properties (e.g., prediction threshold, custom weight function, etc.), which restrict the graph by removing edges pointing to unfeasible counterfactuals. \face then uses the Shortest Path First Algorithm (Dijkstra’s algorithm) over all the instances matching the requirements in the graph. 

We use \face based on the proposed $k$-NN construction algorithm. Since \face provides \textit{counterfactual examples} and not sequences, we then return the lowest-cost counterfactual by applying all the changes sequentially. The action order is randomly chosen. 

\textbf{Hyperparameters.}  CARLA's implementation of \face requires computing an adjacency matrix given the instances in the training set. The procedure is quite expensive since it requires $O(n^2)$ memory, where $n$ is the number of instances. Thus, at inference time, we sample only a fraction of the original dataset. In both \texttt{Adult} and \texttt{GiveMeCredit}, we pick only $10\%$ of the total instances. We set the \textit{number of neighbours}, $k$, to 50 and the distance threshold to $\epsilon = 1.0$ for both datasets. 

\section{Weighted-\fare (\WFARE)}
\label{sec:wfare-full}

\fare is user-agnostic since it is trained by assuming a shared $\vw$ for all instances. Moreover, its reward function optimizes only for the length of the intervention, without explicating the cost $C(I \mid \vw)$. In our experiments, we train \fare with $\bbE_{P(\vw)}[\vw]$. Thus, we push \fare to learn a general policy which works best in the average case, rather than for a specific user. 
To overcome this limitation, \WFARE proposes a new reward function which optimizes for the optimal intervention and tailors its answers to a given set of preferences $\vw$. The new reward is computed as
\begin{equation}
    r = \begin{cases}
    \gamma^{|I| + C(I | \vw)} \cdot \sigma\left(C(I_{\bbE_{P(\vw)}[\vw]} \mid \vw)-C(I \mid \vw)\right) \quad& h(\vs^{(0)}) \neq h(I(\vs^{(0)})) \\
    -1 \quad& \hbox{otherwise}
    \end{cases}
\end{equation}
where $\sigma(x) = \frac{1}{1+ \exp(-x\eta)}$ and $\eta = 0.01$ and $\gamma = 0.97$. The first term $\gamma^{|I| + C(I | \vw)}$ optimizes for short and cheap interventions, while the second term $\sigma(C(I_{\bbE_{P(\vw)}[\vw]} \mid \vw)-C(I \mid \vw))$ encourages \WFARE to provide \textit{personalized} recourse options, by generating sequences which have lower costs than the intervention found by assuming the expected value $\bbE_{P(\vw)}[\vw]$.
We also set the $L(\vs^{(t)}, a')$ term of the \mcts's P-UCT criterion to $c_{act\_cost}\cdot \gamma^{|I| + C(I | \vw)} \cdot \sigma(C(I_{\bbE_{P(\vw)}[\vw]} \mid \vw)-C(I \mid \vw))$ to help the model converge.

In both \rl and \fare, the user preferences are not visible in the state $\vs$. Given a user state $\vs^{(t)}$, \WFARE adds the cost of every valid action $a \in \calA$ as an additional feature. Since an action $a=(f,x)$ is a combination of a function $f$ and an argument $x$, we combine actions relating to the same function into a single feature, averaging their cost over the set of argument values. This simplification allows adding $ \mid \mathcal{F} \mid $ additional features to the user state, where $\mathcal{F}$ is the set of all possible functions. More formally, the user's state $\vs'$ becomes
 \begin{equation}
     \textstyle
     \vs' = \vs \circ \left\{ \frac{1}{ \mid \mathcal{X}_f \mid }\sum_{x \in \mathcal{X}_f}C((f, x),\vs \mid \textbf{w}) \right\} _{f \in \mathcal{F}}
 \end{equation}
where $\mathcal{X}_f$ is the set of all the possible arguments for function $f$. We also considered adding directly the weights $\textbf{w}$, but we discovered that such representation makes it harder to learn an optimal policy.
Ultimately, we train \WFARE by assigning to each user in the training set a dummy weight vector sampled from the prior $\vw \sim P(\vw)$.

\begin{table}[]
\centering
\caption{Performance of all competitors averaged over $10$ runs.  A `-' indicates that the method did not find \textit{any} successful intervention for \textit{any} user. $\mathtt{PEAR}_{NL}$ and $\mathtt{PEAR}_{L}$ indicate \texttt{PEAR} associated with the noiseless and logistic response model, respectively. Please note that while some competitors can sometimes achieve a slightly lower cost compared to $\mathtt{PEAR}$, they suffer from much worse validity.}
\label{tab:eval-results-shallow}
\subcaption*{\textbf{Logistic Regression}}
\resizebox{\textwidth}{!}{%
\begin{tabular}{cccccccc}
\hline
 &  & \multicolumn{3}{c}{\texttt{Adult}} & \multicolumn{3}{c}{\texttt{GiveMeSomeCredit}} \\
\multirow{-2}{*}{\textbf{Users}} & \multirow{-2}{*}{\textbf{Corruption}} & \textit{Validity} & \textit{Cost} & \textit{Length} & \textit{Validity} & \textit{Cost} & \textit{Length} \\ \hline
 & $\mathtt{FARE}$ & \cellcolor[HTML]{EFEFEF}\small{$0.97 \pm 0.16$} & \cellcolor[HTML]{EFEFEF}\small{$266.48 \pm 166.07$} & \cellcolor[HTML]{EFEFEF}\small{$3.44 \pm 0.87$} & \cellcolor[HTML]{EFEFEF}\small{$0.89 \pm 0.18$} & \cellcolor[HTML]{EFEFEF}\small{$156.40 \pm 88.92$} & \cellcolor[HTML]{EFEFEF}\small{$3.15 \pm 0.92$} \\
 & $\mathtt{EFARE}$ & \small{$0.79 \pm 0.35$} & \small{$296.51 \pm 182.87$} & \small{$3.59 \pm 0.98$} & \small{$0.72 \pm 0.34$} & \small{$161.22 \pm 90.93$} & \small{$3.18 \pm 0.95$} \\
 & $\mathtt{RL}$ & \cellcolor[HTML]{EFEFEF}\small{$0.78 \pm 0.37$} & \cellcolor[HTML]{EFEFEF}\small{$268.07 \pm 170.85$} & \cellcolor[HTML]{EFEFEF}\small{$3.43 \pm 0.94$} & \cellcolor[HTML]{EFEFEF}\small{$0.15 \pm 0.36$} & \cellcolor[HTML]{EFEFEF}\small{$58.43 \pm 8.93$} & \cellcolor[HTML]{EFEFEF}\small{$2.00 \pm 0.00$} \\
 & $\mathtt{CSCF}$ & \small{$0.73 \pm 0.26$} & \small{$167.41 \pm 94.55$} & \small{$2.81 \pm 0.44$} & \small{$0.61 \pm 0.39$} & \small{$109.48 \pm 113.21$} & \small{$2.54 \pm 1.04$} \\
 & $\mathtt{MCTS}$ & \cellcolor[HTML]{EFEFEF}\small{$0.45 \pm 0.44$} & \cellcolor[HTML]{EFEFEF}\small{$435.23 \pm 206.31$} & \cellcolor[HTML]{EFEFEF}\small{$4.55 \pm 1.17$} & \cellcolor[HTML]{EFEFEF}\small{$0.72 \pm 0.41$} & \cellcolor[HTML]{EFEFEF}\small{$214.38 \pm 109.90$} & \cellcolor[HTML]{EFEFEF}\small{$4.02 \pm 1.22$} \\
 & $\mathtt{FACE}$ & \small{$0.21 \pm 0.20$} & \small{$419.92 \pm 97.07$} & \small{$3.92 \pm 0.45$} & \small{$0.37 \pm 0.36$} & \small{$319.30 \pm 48.95$} & \small{$5.79 \pm 0.49$} \\
 & $\mathtt{PEAR}_{NL}$ \textbf{(ours)} & \cellcolor[HTML]{EFEFEF}\small{$\bm{0.99 \pm 0.09}$} & \cellcolor[HTML]{EFEFEF}\small{$\bm{146.28 \pm 64.08}$} & \cellcolor[HTML]{EFEFEF}\small{$\bm{2.96 \pm 0.31}$} & \cellcolor[HTML]{EFEFEF}\small{$\bm{0.95 \pm 0.07}$} & \cellcolor[HTML]{EFEFEF}\small{$\bm{90.07 \pm 22.64}$} & \cellcolor[HTML]{EFEFEF}\small{$\bm{2.50 \pm 0.25}$} \\
\multirow{-8}{*}{\texttt{All}} & $\mathtt{PEAR}_{L}$ \textbf{(ours)} & \small{$\bm{0.99 \pm 0.09}$} & \small{$\bm{154.29 \pm 66.55}$} & \small{$\bm{2.98 \pm 0.38}$} & \small{$\bm{0.95 \pm 0.08}$} & \small{$\bm{90.88 \pm 28.86}$} & \small{$\bm{2.49 \pm 0.28}$} \\ \hline
 & $\mathtt{FARE}$ & \small{$0.90 \pm 0.28$} & \small{$471.54 \pm 167.12$} & \small{$4.83 \pm 0.97$} & \small{$0.46 \pm 0.32$} & \small{$332.35 \pm 81.30$} & \small{$5.02 \pm 0.75$} \\
 & $\mathtt{EFARE}$ & \cellcolor[HTML]{EFEFEF}\small{$0.59 \pm 0.45$} & \cellcolor[HTML]{EFEFEF}\small{$484.60 \pm 171.43$} & \cellcolor[HTML]{EFEFEF}\small{$4.84 \pm 0.96$} & \cellcolor[HTML]{EFEFEF}\small{$0.25 \pm 0.36$} & \cellcolor[HTML]{EFEFEF}\small{$346.12 \pm 92.08$} & \cellcolor[HTML]{EFEFEF}\small{$5.16 \pm 0.82$} \\
 & $\mathtt{RL}$ & \small{$0.56 \pm 0.47$} & \small{$466.73 \pm 175.00$} & \small{$4.74 \pm 1.04$} & \small{$0.00 \pm 0.05$} & \small{$83.53 \pm 0.00$} & \small{$2.00 \pm 0.00$} \\
 & $\mathtt{CSCF}$ & \cellcolor[HTML]{EFEFEF}\small{$0.22 \pm 0.34$} & \cellcolor[HTML]{EFEFEF}\small{$395.89 \pm 124.63$} & \cellcolor[HTML]{EFEFEF}\small{$4.20 \pm 0.67$} & \cellcolor[HTML]{EFEFEF}\small{$0.12 \pm 0.30$} & \cellcolor[HTML]{EFEFEF}\small{$190.70 \pm 110.23$} & \cellcolor[HTML]{EFEFEF}\small{$3.40 \pm 0.94$} \\
 & $\mathtt{MCTS}$ & \small{$0.20 \pm 0.38$} & \small{$583.91 \pm 119.72$} & \small{$5.56 \pm 0.59$} & \small{$0.40 \pm 0.43$} & \small{$348.86 \pm 94.76$} & \small{$5.36 \pm 0.94$} \\
 & $\mathtt{FACE}$ & \cellcolor[HTML]{EFEFEF}\small{$0.02 \pm 0.10$} & \cellcolor[HTML]{EFEFEF}\small{$598.63 \pm 11.96$} & \cellcolor[HTML]{EFEFEF}\small{$4.88 \pm 0.15$} & \cellcolor[HTML]{EFEFEF}\small{$0.29 \pm 0.36$} & \cellcolor[HTML]{EFEFEF}\small{$448.70 \pm 39.45$} & \cellcolor[HTML]{EFEFEF}\small{$6.96 \pm 0.49$} \\
 & $\mathtt{PEAR}_{NL}$ \textbf{(ours)} & \small{$\bm{0.95 \pm 0.16}$} & \small{$\bm{320.97 \pm 89.93}$} & \small{$\bm{4.23 \pm 0.36}$} & \small{$\bm{0.73 \pm 0.16}$} & \small{$\bm{259.74 \pm 45.29}$} & \small{$\bm{4.28 \pm 0.46}$} \\
\multirow{-8}{*}{\texttt{Hard}} & $\mathtt{PEAR}_{L}$ \textbf{(ours)} & \cellcolor[HTML]{EFEFEF}\small{$\bm{0.96 \pm 0.15}$} & \cellcolor[HTML]{EFEFEF}\small{$\bm{329.40 \pm 93.83}$} & \cellcolor[HTML]{EFEFEF}\small{$\bm{4.27 \pm 0.42}$} & \cellcolor[HTML]{EFEFEF}\small{$\bm{0.74 \pm 0.17}$} & \cellcolor[HTML]{EFEFEF}\small{$\bm{261.38 \pm 45.67}$} & \cellcolor[HTML]{EFEFEF}\small{$\bm{4.28 \pm 0.47}$} \\ \hline
\end{tabular}%
}
\bigskip
\subcaption*{\textbf{Decision Tree}}
\resizebox{\textwidth}{!}{%
\begin{tabular}{cccccccc}
\hline
 &  & \multicolumn{3}{c}{\texttt{Adult}} & \multicolumn{3}{c}{\texttt{GiveMeSomeCredit}} \\
\multirow{-2}{*}{\textbf{Users}} & \multirow{-2}{*}{\textbf{Corruption}} & \textit{Validity} & \textit{Cost} & \textit{Length} & \textit{Validity} & \textit{Cost} & \textit{Length} \\ \hline
 & $\mathtt{FARE}$ & \cellcolor[HTML]{EFEFEF}\small{$\bm{0.94 \pm 0.15}$} & \cellcolor[HTML]{EFEFEF}\small{$\bm{142.19 \pm 80.14}$} & \cellcolor[HTML]{EFEFEF}\small{$\bm{2.33 \pm 0.39}$} & \cellcolor[HTML]{EFEFEF}\small{$0.67 \pm 0.40$} & \cellcolor[HTML]{EFEFEF}\small{$129.17 \pm 148.29$} & \cellcolor[HTML]{EFEFEF}\small{$2.70 \pm 1.18$} \\
 & $\mathtt{EFARE}$ & \small{$0.71 \pm 0.22$} & \small{$261.22 \pm 188.57$} & \small{$3.18 \pm 0.98$} & \small{$0.81 \pm 0.31$} & \small{$213.55 \pm 105.56$} & \small{$3.93 \pm 1.09$} \\
 & $\mathtt{RL}$ & \cellcolor[HTML]{EFEFEF}\small{$0.70 \pm 0.31$} & \cellcolor[HTML]{EFEFEF}\small{$288.06 \pm 198.03$} & \cellcolor[HTML]{EFEFEF}\small{$3.37 \pm 1.13$} & \cellcolor[HTML]{EFEFEF}\small{$0.64 \pm 0.39$} & \cellcolor[HTML]{EFEFEF}\small{$235.94 \pm 109.32$} & \cellcolor[HTML]{EFEFEF}\small{$4.14 \pm 1.16$} \\
 & $\mathtt{CSCF}$ & \small{$0.55 \pm 0.31$} & \small{$294.45 \pm 200.01$} & \small{$3.43 \pm 1.07$} & \small{$0.07 \pm 0.25$} & \small{$96.50 \pm 105.43$} & \small{$2.65 \pm 0.81$} \\
 & $\mathtt{MCTS}$ & \cellcolor[HTML]{EFEFEF}\small{$0.47 \pm 0.46$} & \cellcolor[HTML]{EFEFEF}\small{$452.30 \pm 225.37$} & \cellcolor[HTML]{EFEFEF}\small{$4.50 \pm 1.31$} & \cellcolor[HTML]{EFEFEF}\small{$0.73 \pm 0.42$} & \cellcolor[HTML]{EFEFEF}\small{$229.19 \pm 122.03$} & \cellcolor[HTML]{EFEFEF}\small{$4.13 \pm 1.32$} \\
 & $\mathtt{FACE}$ & \small{$0.23 \pm 0.25$} & \small{$450.83 \pm 112.93$} & \small{$4.05 \pm 0.57$} & \small{$0.59 \pm 0.39$} & \small{$328.81 \pm 55.37$} & \small{$5.72 \pm 0.58$} \\
 & $\mathtt{PEAR}_{NL}$ \textbf{(ours)} & \cellcolor[HTML]{EFEFEF}\small{$0.88 \pm 0.14$} & \cellcolor[HTML]{EFEFEF}\small{$169.03 \pm 64.91$} & \cellcolor[HTML]{EFEFEF}\small{$2.64 \pm 0.36$} & \cellcolor[HTML]{EFEFEF}\small{$\bm{0.95 \pm 0.11}$} & \cellcolor[HTML]{EFEFEF}\small{$\bm{113.21 \pm 29.33}$} & \cellcolor[HTML]{EFEFEF}\small{$\bm{2.70 \pm 0.28}$} \\
\multirow{-8}{*}{\texttt{All}} & $\mathtt{PEAR}_{L}$ \textbf{(ours)} & \small{$0.88 \pm 0.13$} & \small{$168.41 \pm 57.40$} & \small{$2.65 \pm 0.38$} & \small{$\bm{0.95 \pm 0.11}$} & \small{$\bm{113.47 \pm 31.94}$} & \small{$\bm{2.69 \pm 0.31}$} \\ \hline
 & $\mathtt{FARE}$ & \small{$\bm{0.92 \pm 0.14}$} & \small{$\bm{117.19 \pm 54.11}$} & \small{$\bm{2.32 \pm 0.34}$} & \small{$0.44 \pm 0.43$} & \small{$171.40 \pm 153.66$} & \small{$3.09 \pm 1.27$} \\
 & $\mathtt{EFARE}$ & \cellcolor[HTML]{EFEFEF}\small{$0.80 \pm 0.18$} & \cellcolor[HTML]{EFEFEF}\small{$226.16 \pm 191.83$} & \cellcolor[HTML]{EFEFEF}\small{$3.12 \pm 1.05$} & \cellcolor[HTML]{EFEFEF}\small{$0.56 \pm 0.39$} & \cellcolor[HTML]{EFEFEF}\small{$270.40 \pm 126.42$} & \cellcolor[HTML]{EFEFEF}\small{$4.44 \pm 1.16$} \\
 & $\mathtt{RL}$ & \small{$0.76 \pm 0.24$} & \small{$233.65 \pm 197.26$} & \small{$3.19 \pm 1.06$} & \small{$0.35 \pm 0.41$} & \small{$318.28 \pm 132.66$} & \small{$4.81 \pm 1.13$} \\
 & $\mathtt{CSCF}$ & \cellcolor[HTML]{EFEFEF}\small{$0.62 \pm 0.29$} & \cellcolor[HTML]{EFEFEF}\small{$229.10 \pm 198.01$} & \cellcolor[HTML]{EFEFEF}\small{$3.07 \pm 1.04$} & \cellcolor[HTML]{EFEFEF}\small{$0.02 \pm 0.15$} & \cellcolor[HTML]{EFEFEF}\small{$129.69 \pm 23.40$} & \cellcolor[HTML]{EFEFEF}\small{$2.68 \pm 0.45$} \\
 & $\mathtt{MCTS}$ & \small{$0.44 \pm 0.46$} & \small{$424.77 \pm 228.23$} & \small{$4.52 \pm 1.31$} & \small{$0.57 \pm 0.45$} & \small{$300.49 \pm 134.82$} & \small{$4.77 \pm 1.27$} \\
 & $\mathtt{FACE}$ & \cellcolor[HTML]{EFEFEF}\small{$0.08 \pm 0.20$} & \cellcolor[HTML]{EFEFEF}\small{$463.89 \pm 86.34$} & \cellcolor[HTML]{EFEFEF}\small{$4.36 \pm 0.51$} & \cellcolor[HTML]{EFEFEF}\small{$0.49 \pm 0.41$} & \cellcolor[HTML]{EFEFEF}\small{$457.94 \pm 57.34$} & \cellcolor[HTML]{EFEFEF}\small{$6.65 \pm 0.58$} \\
 & $\mathtt{PEAR}_{NL}$ \textbf{(ours)} & \small{$0.84 \pm 0.12$} & \small{$123.20 \pm 44.21$} & \small{$2.50 \pm 0.30$} & \small{$\bm{0.83 \pm 0.21}$} & \small{$\bm{174.23 \pm 41.69}$} & \small{$\bm{3.44 \pm 0.40}$} \\
\multirow{-8}{*}{\texttt{Hard}} & $\mathtt{PEAR}_{L}$ \textbf{(ours)} & \cellcolor[HTML]{EFEFEF}\small{$0.84 \pm 0.12$} & \cellcolor[HTML]{EFEFEF}\small{$124.37 \pm 50.78$} & \cellcolor[HTML]{EFEFEF}\small{$2.50 \pm 0.33$} & \cellcolor[HTML]{EFEFEF}\small{$\bm{0.84 \pm 0.20}$} & \cellcolor[HTML]{EFEFEF}\small{$\bm{175.30 \pm 42.79}$} & \cellcolor[HTML]{EFEFEF}\small{$\bm{3.47 \pm 0.44}$} \\ \hline
\end{tabular}%
}
\end{table}

\section{Evaluating \method on shallow classifiers}
\label{sec:pear-shallow-classifiers}

We evaluated the performances of \method also by using non-neural machine learning models. Namely, we trained a logistic regression and a tree-based classifier. We performed hyperparameter optimization using grid search \eg the \textit{splitting criterion} (decision tree) and strength for the regularizer (logistic model). We then replicated the experiments described in \cref{sec:experiments} following the same protocol. In the case of the decision tree, the score of the classifier $h(\vx)$ is represented by the fraction of the examples of the same class in a leaf.
\cref{tab:eval-results-shallow} shows the evaluation results. \method tends to outperform the competitors in terms of cost and validity in 6 out of 6 (\texttt{GiveMeSomeCredit}) and 4 out of 6 (\texttt{Adult}) experiments. The only exception is decision trees on \texttt{Adult}, where \method is the runner-up behind \texttt{CSCF}, and still outperforms all other competitors.

\begin{figure}[t]
    \centering
    \includegraphics[width=0.60\linewidth]{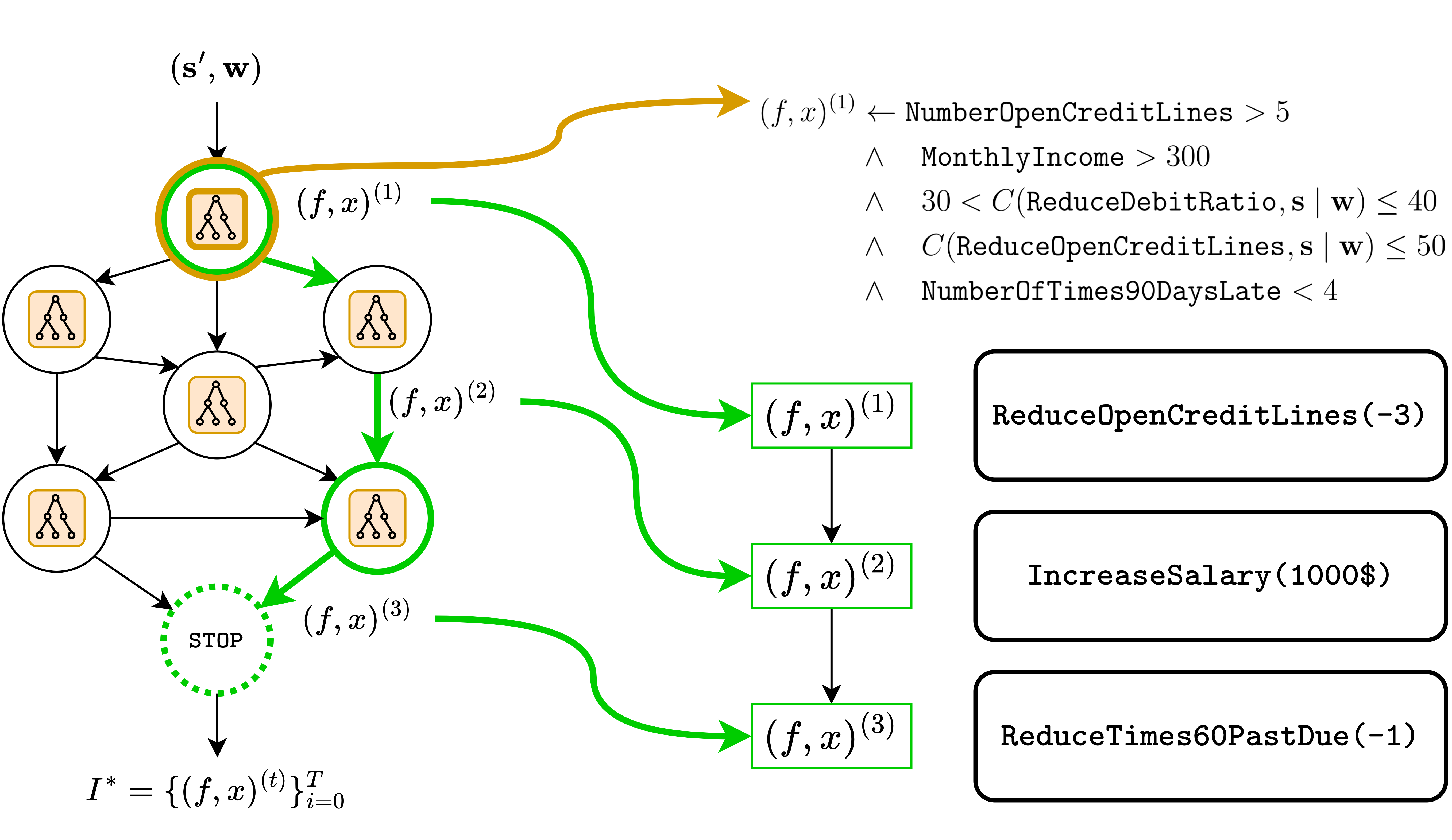}
    \caption{\textbf{The \texttt{W-EFARE} method.} Given $(\vs', \vw)$, we traverse the graph (green path) using the decision tree (orange components) to choose the next action $(f,x)^{(t)}$ from the available transitions. When we reach the \texttt{STOP} node, we return the found intervention $I^*$. On the right, we have a generated intervention achieving recourse for the \texttt{GiveMeSomeCredit} dataset. On the top-right, we have the Boolean rule extracted from the decision tree motivating the first recommended action $\mathtt{ReduceOpenCreditLines}$. As we can see, \texttt{W-EFARE} can generate rules that also depend on the cost function $C$, achieving the desired user awareness.}
    \label{fig:w-efare}
\end{figure}

\section{Explaining Recommended Interventions with \xmethod}
\label{sec:user-aware-explainable-interventions}

\method asks the users to choose the best option from a small set of interventions. The task of picking the best intervention requires cognitive effort since the user must understand all the recommended sequences of actions. This can be problematic if the user does not comprehend the reason behind the recommendations, undermining the quality of the feedback they provide and their overall trust in the system. To overcome this problem, we additionally developed a variant of \method that can complement interventions with action-specific explanations.

As detailed in \cref{sec:fare-efare}, \EFARE \citep{detoni2022synthesizing} is a deterministic user-agnostic model for generating recourse by providing Boolean explanations for each action in the intervention. 
The procedure can be adapted to account for user-specific costs. The resulting model, which we call \texttt{W-EFARE}, is a user-aware version of \texttt{EFARE} that can provide \textit{personalized} explanations complementing the interventions. 
\cref{fig:w-efare} shows a graphical illustration of how \texttt{W-EFARE} works in practice.
We use this method to develop \xmethod, an explainable version of \method, which uses \WEFARE to generate both interventions and explanations which can be shown to the user in the choice set $O$. 

\cref{tab:xpear-results} shows the evaluation results for \xmethod. We compare it against \method, \fare and \EFARE since they share similar characteristics and assumptions. For the \texttt{Adult} dataset, \xmethod provides cheaper recourse options than the competitors (\fare and \EFARE) by keeping a comparable or better validity. The behaviour is consistent for both \texttt{All} and \texttt{Hard} users. For the \texttt{GiveMeSomeCredit} dataset, \xmethod generates sequences with a better or similar validity and cost than the competitors, although the effect is diminished. In both datasets, \xmethod has higher validity than its deterministic counterpart \EFARE.  Overall, \method maintains the upper hand. However, it is worth noticing that \xmethod provides additional feedback to the users, which might be a valuable addition during the elicitation process.

\section{Further evaluations of \method}
\label{sec:appendix-further-evaluations}

\begin{figure}
    \centering
    \begin{subfigure}{\linewidth}
        \includegraphics[width=\linewidth]{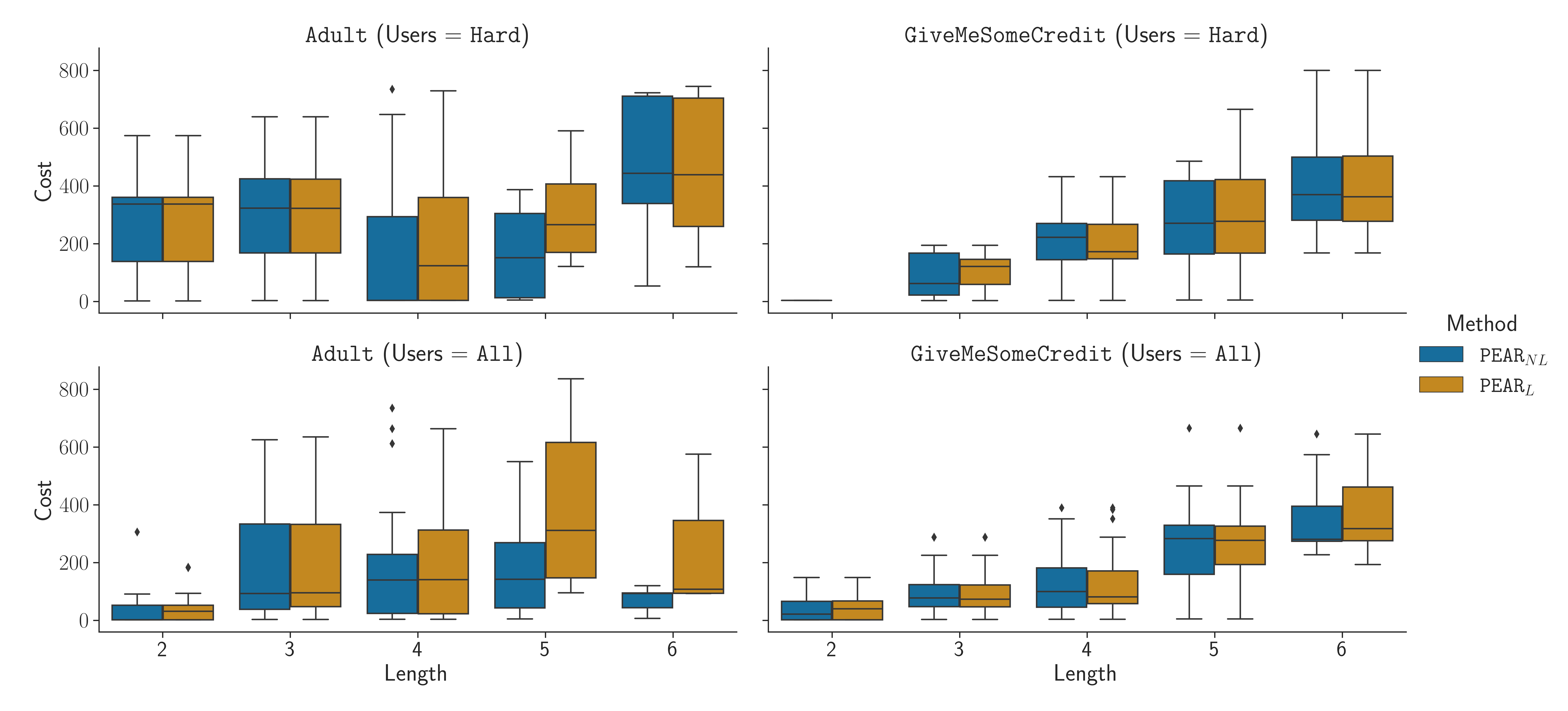}
        \caption{$\mathtt{PEAR}_{NL}$ and $\mathtt{PEAR}_{L}$.}
    \end{subfigure}\vfill
    \begin{subfigure}{\linewidth}
        \includegraphics[width=\linewidth]{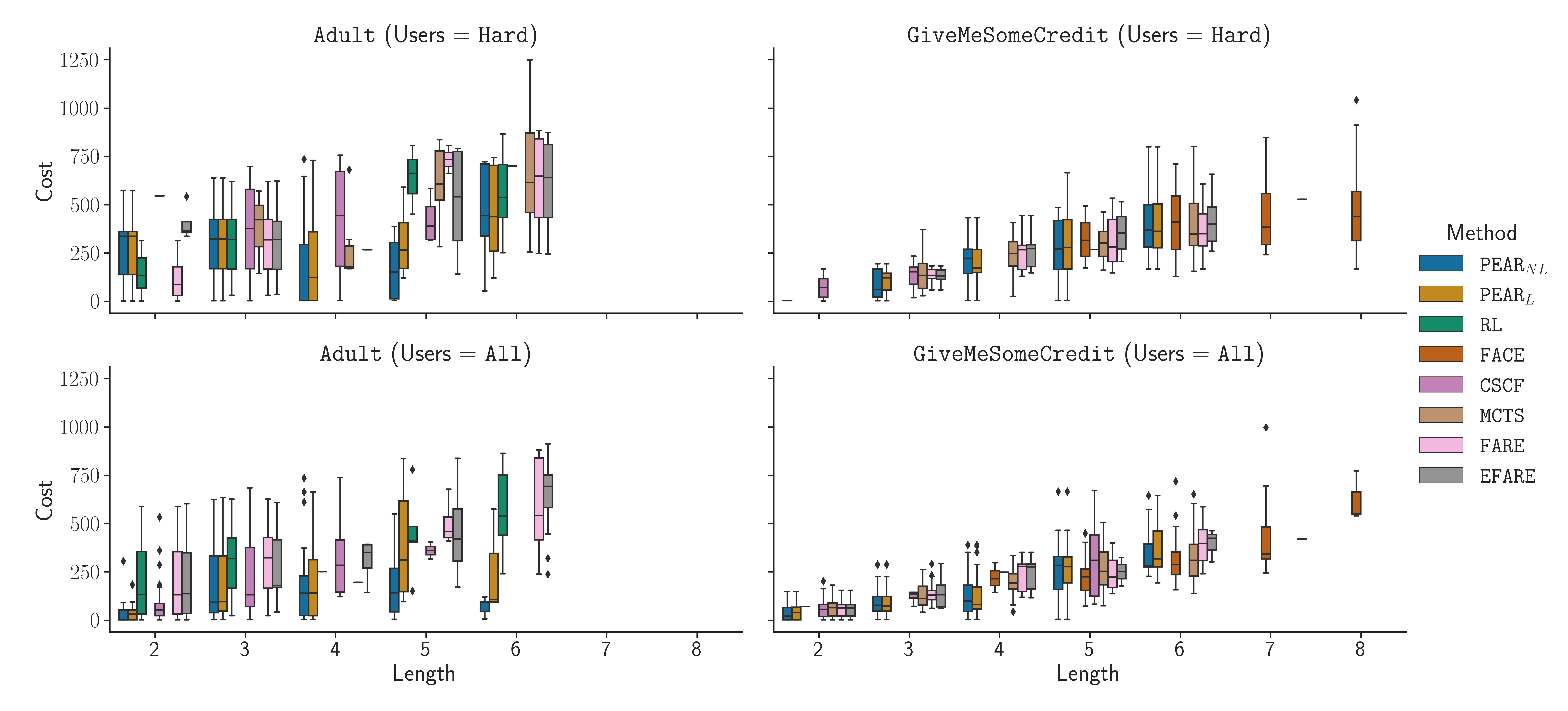}
        \caption{All the methods.}
    \end{subfigure}
    \caption{\textbf{Cost vs Intervention Length.} (Top) In the \texttt{Adult} dataset, \method generates interventions whose costs do not depend on their length (for both \texttt{All} and \texttt{Hard} users). The relationship is more pronounced on the \texttt{GiveMeSomeCredit} dataset, but longer interventions (e.g., $|I| = 5$) still present high variability in the costs. These results hint at the fact that length alone can be a bad proxy to measure the "effort" of a user. (Bottom) We plot the same results for all the methods. The relationship between cost and length is well represented here, and it aligns with the previous findings.}
    \label{fig:cost-vs-length}
\end{figure}

\begin{figure}
    \centering
    \begin{subfigure}{\linewidth}
        \includegraphics[width=\textwidth]{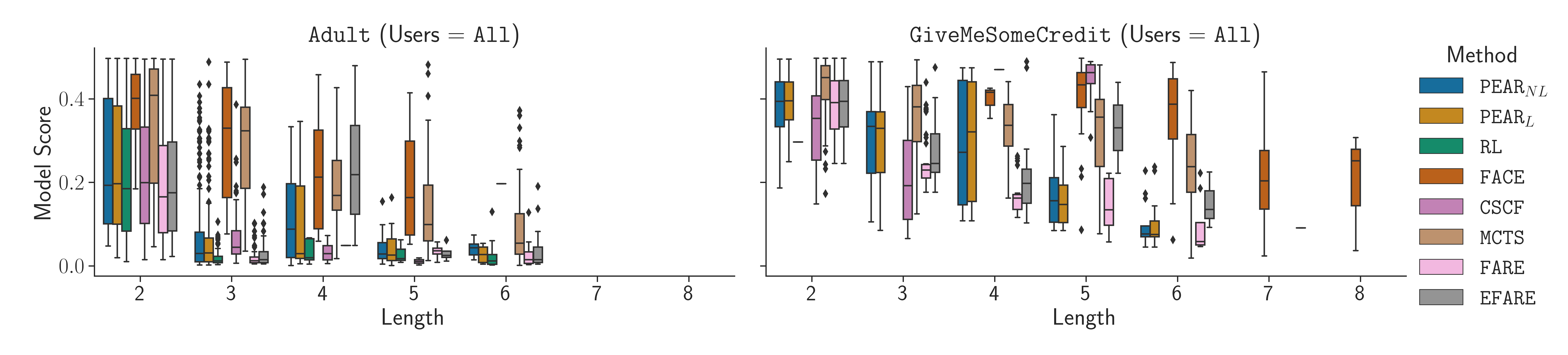}
        \caption{Model Score vs Length.}
    \end{subfigure}\vfill%
    \begin{subfigure}{\linewidth}
        \includegraphics[width=\linewidth]{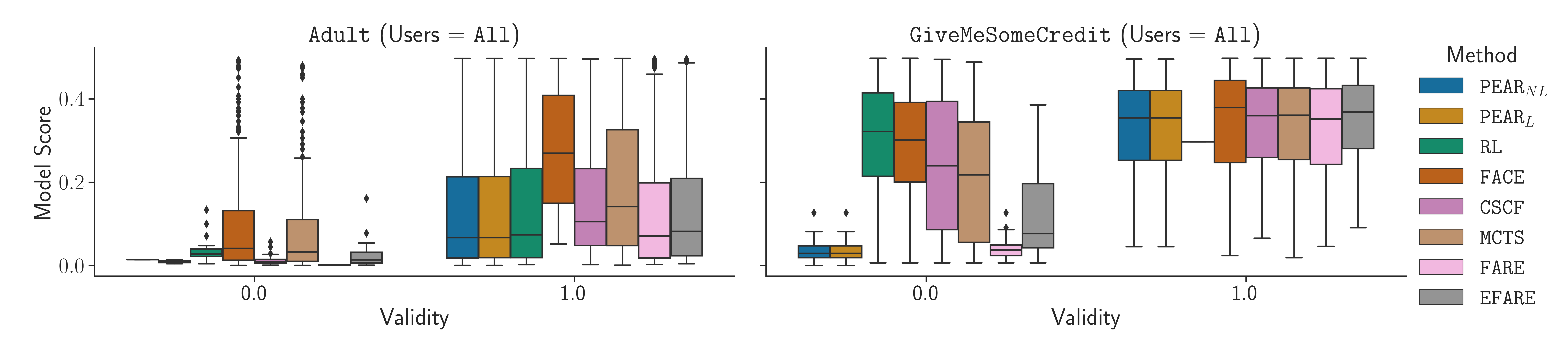}
        \caption{Model Score vs Validity.}
    \end{subfigure}
    \caption{\textbf{Model Score vs Length (Top).} In the \texttt{Adult} dataset there is a clear correlation between the score and the length of the interventions. Lower scores indicate harder individuals who need longer sequences to obtain recourse. In the \texttt{GiveMeSomeCredit} case, we can still see the trend, even if \face and \mcts seem to be less susceptible. \textbf{Model Score vs Length (Validity) (bottom).} In the \texttt{Adult} experiment, all the models have low validity only in the lower-scoring users. In the \texttt{GiveMeSomeCredit} experiments, \method and \fare are the only two methods which have low validity only for the \texttt{Hard} users, while the others tend to fail more often, irrespectively of the classifier score.}
    \label{fig:score-vs-length}
\end{figure}

We now present some graphs showing the relationship between the cost and length of the interventions (\cref{fig:cost-vs-length}) and between the black box model score and the validity/interventions length (\cref{fig:score-vs-length}).

In \cref{fig:cost-vs-length}, we can see that the length of an intervention can be a bad proxy for the intervention's difficulty. Namely, there are long interventions which still retain a small cost or effort for the user. Thus, we need to optimize for both the length and cost of interventions simultaneously. This notion is reflected in the reward function of \WFARE as we mentioned in \cref{sec:wfare-full}.

In \cref{fig:score-vs-length}, we see how the black box prediction score is a good metric to understand how hard it is for a user to obtain recourse. Namely, we suggest longer interventions for users obtaining a low score with respect to the others. Moreover, all the methods tend to fail to provide recourse to users with low scores. Interestingly, methods which learn a recourse policy rather than performing direct optimization (e.g., \method and \fare) are more capable of providing recourse generalizing over the low-score users. 

\begin{table}
\centering
\caption{Performance of \xmethod averaged over $10$ runs. $\mathtt{XPEAR}_{NL}$ and $\mathtt{XPEAR}_{L}$ indicate \xmethod associated with the noiseless and logistic response model, respectively. The best results are boldfaced.}
\label{tab:xpear-results}
\resizebox{\textwidth}{!}{%
\begin{tabular}{@{}cccccccc@{}}
\toprule
 &  & \multicolumn{3}{c}{\texttt{Adult}} & \multicolumn{3}{c}{\texttt{GiveMeSomeCredit}} \\ \cmidrule(l){3-8} 
\multirow{-2}{*}{\textbf{Users}} & \multirow{-2}{*}{\textbf{Corruption}} & \textit{Validity} & \textit{Cost} & \textit{Length} & \textit{Validity} & \textit{Cost} & \textit{Length} \\ \midrule
 & \cellcolor[HTML]{EFEFEF}$\mathtt{FARE}$ & \cellcolor[HTML]{EFEFEF}\small{$0.90 \pm 0.27$} & \cellcolor[HTML]{EFEFEF}\small{$285.63 \pm 195.68$} & \cellcolor[HTML]{EFEFEF}\small{$3.45 \pm 1.19$} & \cellcolor[HTML]{EFEFEF}\small{$0.86 \pm 0.23$} & \cellcolor[HTML]{EFEFEF}\small{$161.77 \pm 107.42$} & \cellcolor[HTML]{EFEFEF}\small{$3.27 \pm 1.13$} \\
 & $\mathtt{EFARE}$ & \small{$0.76 \pm 0.39$} & \small{$306.11 \pm 199.02$} & \small{$3.54 \pm 1.25$} & \small{$0.67 \pm 0.38$} & \small{$155.92 \pm 109.19$} & \small{$3.18 \pm 1.18$} \\
 & \cellcolor[HTML]{EFEFEF}$\mathtt{PEAR}_{NL}$ & \cellcolor[HTML]{EFEFEF}\small{$\bm{1.00 \pm 0.03}$} & \cellcolor[HTML]{EFEFEF}\small{$\bm{142.23 \pm 61.75}$} & \cellcolor[HTML]{EFEFEF}\small{$\bm{2.84 \pm 0.59}$} & \cellcolor[HTML]{EFEFEF}\small{$\bm{0.89 \pm 0.00}$} & \cellcolor[HTML]{EFEFEF}\small{$\bm{96.04 \pm 31.96}$} & \cellcolor[HTML]{EFEFEF}\small{$\bm{2.79 \pm 0.42}$} \\
 & $\mathtt{PEAR}_{L}$ & \small{$\bm{1.00 \pm 0.04}$} & \small{$\bm{146.54 \pm 63.09}$} & \small{$\bm{2.84 \pm 0.56}$} & \small{$\bm{0.89 \pm 0.00}$} & \small{$\bm{100.19 \pm 29.53}$} & \small{$\bm{2.85 \pm 0.48}$} \\
 & \cellcolor[HTML]{EFEFEF}$\mathtt{XPEAR}_{NL}$ & \cellcolor[HTML]{EFEFEF}\small{$0.97 \pm 0.12$} & \cellcolor[HTML]{EFEFEF}\small{$183.55 \pm 97.10$} & \cellcolor[HTML]{EFEFEF}\small{$3.00 \pm 0.71$} & \cellcolor[HTML]{EFEFEF}\small{$0.83 \pm 0.18$} & \cellcolor[HTML]{EFEFEF}\small{$125.00 \pm 44.88$} & \cellcolor[HTML]{EFEFEF}\small{$2.84 \pm 0.40$} \\
\multirow{-6}{*}{\texttt{All}} & $\mathtt{XPEAR}_{L}$ & \small{$0.97 \pm 0.13$} & \small{$203.08 \pm 112.29$} & \small{$2.99 \pm 0.78$} & \small{$0.84 \pm 0.19$} & \small{$129.59 \pm 51.57$} & \small{$2.85 \pm 0.47$} \\ \midrule
 & $\mathtt{FARE}$ & \small{$0.71 \pm 0.44$} & \small{$438.97 \pm 188.50$} & \small{$4.54 \pm 1.21$} & \small{$0.47 \pm 0.37$} & \cellcolor[HTML]{FFFFFF}\small{$319.46 \pm 96.36$} & \small{$4.97 \pm 0.70$} \\
 & \cellcolor[HTML]{EFEFEF}$\mathtt{EFARE}$ & \cellcolor[HTML]{EFEFEF}\small{$0.55 \pm 0.48$} & \cellcolor[HTML]{EFEFEF}\small{$454.05 \pm 202.76$} & \cellcolor[HTML]{EFEFEF}\small{$4.52 \pm 1.25$} & \cellcolor[HTML]{EFEFEF}\small{$0.22 \pm 0.36$} & \cellcolor[HTML]{EFEFEF}\small{$371.58 \pm 82.18$} & \cellcolor[HTML]{EFEFEF}\small{$5.31 \pm 0.71$} \\
 & $\mathtt{PEAR}_{NL}$ & \small{$\bm{0.99 \pm 0.08}$} & \small{$\bm{296.37 \pm 43.84}$} & \small{$\bm{3.35 \pm 0.55}$} & \small{$\bm{0.58 \pm 0.04}$} & \small{$\bm{251.60 \pm 51.16}$} & \small{$\bm{4.59 \pm 0.40}$} \\
 & \cellcolor[HTML]{EFEFEF}$\mathtt{PEAR}_{L}$ & \cellcolor[HTML]{EFEFEF}\small{$\bm{0.99 \pm 0.09}$} & \cellcolor[HTML]{EFEFEF}\small{$\bm{301.13 \pm 52.61}$} & \cellcolor[HTML]{EFEFEF}\small{$\bm{3.34 \pm 0.58}$} & \cellcolor[HTML]{EFEFEF}\small{$\bm{0.58 \pm 0.02}$} & \cellcolor[HTML]{EFEFEF}\small{$\bm{262.23 \pm 45.36}$} & \cellcolor[HTML]{EFEFEF}\small{$\bm{4.64 \pm 0.36}$} \\
 & $\mathtt{XPEAR}_{NL}$ & \small{$0.96 \pm 0.13$} & \small{$330.91 \pm 99.51$} & \small{$3.49 \pm 0.65$} & \small{$0.44 \pm 0.30$} & \small{$300.97 \pm 62.14$} & \small{$4.81 \pm 0.50$} \\
\multirow{-6}{*}{\texttt{Hard}} & \cellcolor[HTML]{EFEFEF}$\mathtt{XPEAR}_{L}$ & \cellcolor[HTML]{EFEFEF}\small{$0.97 \pm 0.13$} & \cellcolor[HTML]{EFEFEF}\small{$334.58 \pm 96.28$} & \cellcolor[HTML]{EFEFEF}\small{$3.46 \pm 0.66$} & \cellcolor[HTML]{EFEFEF}\small{$0.45 \pm 0.31$} & \cellcolor[HTML]{EFEFEF}\small{$313.95 \pm 66.94$} & \cellcolor[HTML]{EFEFEF}\small{$4.88 \pm 0.56$} \\ \bottomrule
\end{tabular}%
}
\end{table}

\end{document}